%% file: acl_latex.tex
\documentclass[11pt]{article}

\usepackage[final]{acl}

\usepackage{times}
\usepackage{latexsym}

\usepackage[T1]{fontenc}

\usepackage[utf8]{inputenc}

\usepackage{microtype}

\usepackage{inconsolata}

\usepackage{graphicx}

\usepackage{booktabs}
\usepackage{amsmath}

\newcommand{\unknown}[1]{{\color{red} (UNKNOWN)}}

\title{Persona Non Grata: LLM Persona-Driven Generations\\in MCQA are Unstable in Distinct Dimensions}

\author{
 \textbf{César Guerra-Solano},
 \textbf{Xiang Lorraine Li}
\\
 Department of Computer Science, University of Pittsburgh
\\
 \small{
   \textbf{Correspondence:} \texttt{\{cguerrasol, xianglli\}@pitt.edu}
 }
}

\begin{document}
\maketitle

\input{latex/0_sections/0_abstract}
\input{latex/0_sections/1_introduction}
\input{latex/0_sections/2_related_work}
\input{latex/0_sections/3_metrics}
\input{latex/0_sections/4_experiments}
\input{latex/0_sections/5_results_analysis}
\input{latex/0_sections/7_conclusion}
\input{latex/0_sections/limitations_considerations_acknowledgments}

\bibliography{custom}
\input{latex/0_sections/8_appendix}

\end{document}

%% file: latex/0_sections/0_abstract.tex
\begin{abstract}
Persona-driven generations (PDGs) have seen prolific use in research and industry applications, where a large language model (LLM) takes on a “persona” while completing some task. While persona expressed through free-form text (like dialogue) has substantial work investigating stability or consistency, relatively, persona expressed in non-text-heavy outputs (like in multiple-choice question answering, or MCQA) is often overlooked.
We work to address this gap, seeking to understand the instability of LLM PDGs in MCQA tasks. We develop three metrics investigating the performance, outcome, and question correctness stability, evaluating three distinct dimensions. 
Using these metrics, we find that instability varies consistently between model families and model size, and across question domains, with math/commonsense questions leading to greater instability. We also find task prompt format introduces more prediction instability than other hyperparameters, like temperature. Finally, we find that instability is related to task accuracy, and using our instability metrics, find different experimental settings that result in different best and worst personas for tasks, despite their similarity. This reveals the importance of checking hyperparameter instability in PDGs.

\end{abstract}

%% file: latex/0_sections/1_introduction.tex
\section{Introduction}
\label{sec:introduction}

\input{latex/3_figure_tex/demonstration_of_instability}

Large language models (LLMs) have seen prolific use in a variety of domains, with their adaptability and broad capabilities lending themselves to high performance across a range of tasks and opportunities for personalization for user- and task-specific needs~\cite{kojima2022large, brown2020language, grattafiori2024llama, yang2025qwen3, zhang2024personalization}. With this, \textbf{persona-driven generations (PDGs)} have become prevalent -- here, by leveraging the power of prompting, LLMs can role-play as a "persona" while carrying out some task, such as general user assistance (e.g. "You are a teacher. Explain [...]") or more specific task completion (e.g. "You are a hiring manager. Rate these resumes [...]"). As PDGs see use in high-stakes domains, from medicine to education, evaluating their stability is critical~\cite{sun2025mockllm-roleplaying-in-hiring,yuan2025simulating-roleplaying-in-education, kyung2025patientsim-role-playing-in-medicine, li2024leveraging-role-playing-in-medicine}.

With PDGs, two main uses can be observed: in free-form text generation settings, such as in dialogue systems like Character.AI~\cite{characterai}, and non-text-heavy outputs, such as in persona-based multiple-choice question answering (MCQA). In the latter case, LLM persona is expressed through task completion: rather than producing text that reflects the \textit{style} of a persona, an LLM answers questions in accordance with the \textit{capabilities} associated with that persona, when relevant (e.g., a biologist answering biology questions), or to remain unaffected by persona choice when the persona is irrelevant (e.g., race should not influence performance on mathematical questions).

However, relative to text generation, little work has characterized or improved PDGs in MCQA. Prior studies indicate that PDGs in this setting are implicitly biased and unpredictable~\cite{gupta2024personabias, zheng-jurgens-etal-2024-helpful}, motivating further work to characterize them and their potential flaws. Additionally, these studies bear little standardization, with large differences in model choice and hyperparameters across their experiment settings -- features known to affect model performance, thus being targets of optimization~\cite{wang2023cost, sclar2023quantifying-spurious-prompt-features}. With the diverse experiment settings that PDGs are used in, differing in persona, task, and prompt format, \textbf{additional investigations are necessary to find the effects of these variations} to ensure current and future findings are robust.

We work to address this gap, seeking to further understand PDGs in MCQA, quantifying their stability across diverse environments. We do so relative to \textbf{three persona-specific metrics}, targeting unique and distinct features of PDG paradigms -- we look at not only stability of \textbf{accuracy} as in related past work~\cite{gupta2024personabias, zheng-jurgens-etal-2024-helpful, de2025principled}, but also stability of \textbf{experimental outcomes} and \textbf{persona-specific question correctness}.

We evaluate 4 open-source models across 48 experiment settings and 41 personas, finding:
\begin{itemize}
    \item PDGs in MCQA are \textbf{heavily unstable}, with great variation across similar experiment settings used in past work, and consistent patterns with model size and family.
    \item Instability is \textbf{specific}, with distinct relationships relative to 3 dimensions of instability, types of content in the input prompt (such as the question formatting vs. persona formatting), and the subject domain of the task.
    \item Our instability metrics are \textbf{actionable}, with demonstrated usage in identifying stable and unstable experiment settings that greatly differ in overall accuracy and persona-related conclusions, such as the best persona for a task, or inter-persona distribution shifts.
\end{itemize}

%% file: latex/3_figure_tex/demonstration_of_instability.tex
\begin{figure}
    \centering
    \includegraphics[width=\columnwidth]{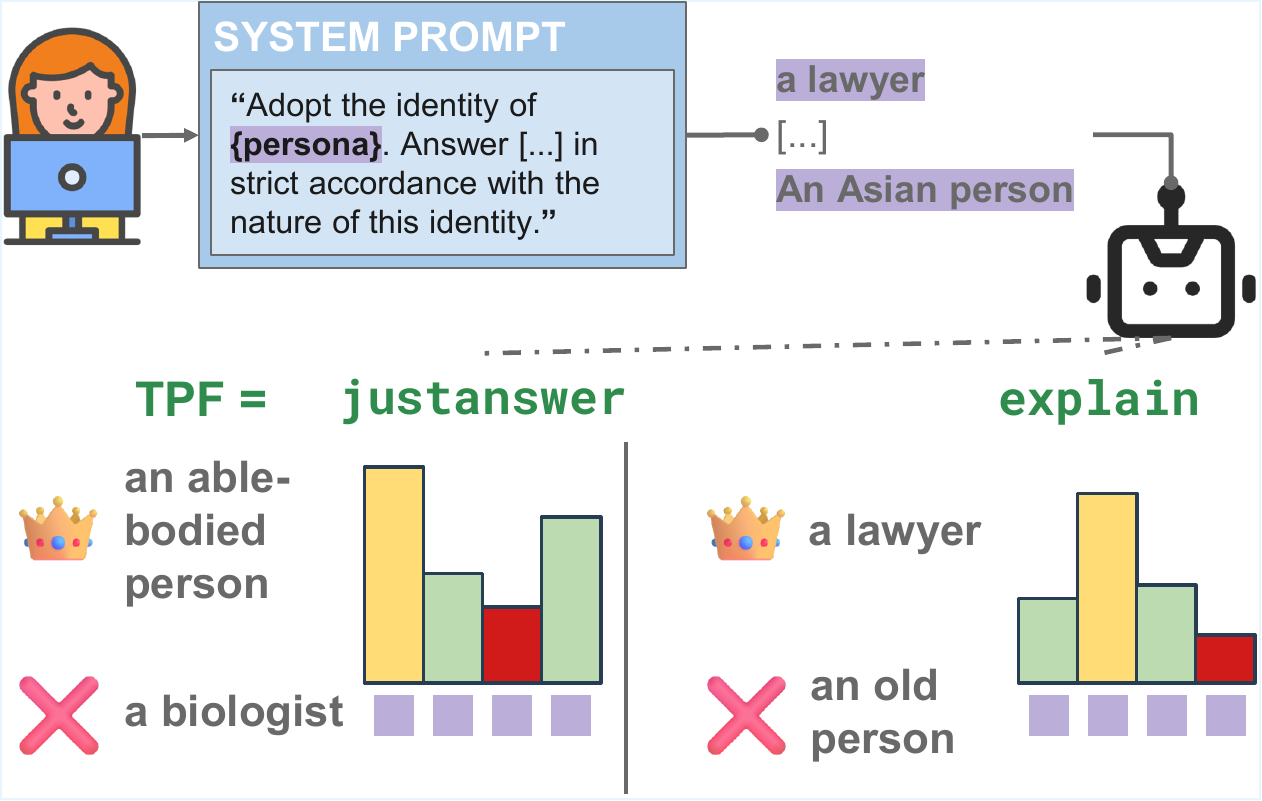}
    \caption{Depicting instability in non-text-heavy persona-driven generations (PDGs). Given different, although still reasonable, experiment configurations, represented by different settings for the task prompt format (TPF) hyperparameter, an LLM's persona-driven performance on a multiple-choice evaluation differs greatly.}
    \label{fig:example_instability}
\end{figure}



%% file: latex/0_sections/2_related_work.tex
\section{Related Work}
\label{sec:related_work}

\noindent\textbf{Persona-Driven Generations} Persona-based generation is widely used across LLM downstream tasks, including dialogue systems~\cite{huang2023personalized}, narrative generation~\cite{zhang2024personalization}, and LLM-as-a-judge settings where models adopt expert roles~\cite{dong-etal-2024-llm, yu2024large, li2025llm}. These applications generally fall into two categories: free-form text generation, where persona is embedded into generation style and content (e.g., Character.AI~\cite{characterai}), and non-text-heavy tasks, like persona-based task-taking, where persona is reflected in task performance.

Evaluation of PDGs has primarily focused on text-heavy settings, examining consistency, stability, and failures such as sensitivity or bias in long-form generation~\cite{zhang-etal-2025-persona, liu2024evaluating, maharana2024evaluating, tseng-etal-2024-two, de2025principled, cheng2023marked}. By contrast, non-text-heavy PDGs—though common in tasks like LLM-as-a-judge or MCQA~\cite{de2025principled}—remain underexplored, with only limited attention to bias, faithfulness, or unpredictability~\cite{zheng-jurgens-etal-2024-helpful, gupta2024personabias, de2025principled}, leaving a key gap in comprehensively assessing their robustness.

Our work is related to~\citet{de2025principled} which evaluates how well personas \textit{match human expectations or needs} with personas, like if expert personas confer a performance advantage. However, we focus on how \textit{stable} PDGs are, i.e, the consistency of MCQA performance given personas across noted variations in experiment setting such as prompt format and generation temperature.

\noindent\textbf{Generation Sensitivity}
Prior work on model generation sensitivity has primarily focused on model temperature and input prompt styles/formatting, with evaluation metrics emphasizing macro-level aspects such as task or response structure and performance~\cite{sclar2023quantifying-spurious-prompt-features, chatterjee2024posix, li2025llm, brown2020language, he2024does}. However, these studies primarily center on persona-agnostic contexts, leaving the need to investigate how experiment setting variations influence LLMs’ role-playing capabilities. Moreover, existing sensitivity metrics~\cite{zhuo-etal-2024-prosa-promptsensiscore} largely focus on accuracy, overlooking outcome- and question-level sensitivity, a gap our work addresses.

%% file: latex/0_sections/3_metrics.tex
\section{Instability Metrics}
\input{latex/3_figure_tex/demonstration_of_metrics}
\label{sec:metric_formulation}

In this paper, we define instability in persona-driven generations as robustness to experimental settings. We examine how evaluation trends shift across hyperparameter settings such as prompt format and model temperature, with each setting $s \in S$ representing a hyperparameter combination.

To quantify performance variation, we propose three metrics that measure differences in \textbf{accuracy}, \textbf{outcome}, and \textbf{correct questions}. Prior work~\cite{zhuo-etal-2024-prosa-promptsensiscore, sclar2023quantifying-spurious-prompt-features} has primarily focused only on accuracy differences.
However, instability exists agnostic of accuracy -- you can have large accuracy differences but no difference in inter-persona ranking, or identical  accuracy but great inconsistency in \textit{what} questions are answered correctly. Thus, further dimensions of instability should be investigated, especially in PDGs. We define each metric using the following terms.\footnote{For validation, we report the correlation of our metric results with metrics from existing literature or reasonable correlates in Appendix~\ref{app:metric_correlations}.}

\noindent\textbf{Experimental Settings:} Each metric is calculated relative to a set of considered experimental settings, $S$. This allows us to analyze instability on broader (e.g., considering all experimental settings) or smaller (e.g., varying one hyperparameter within an otherwise fixed setting) scales.

\noindent\textbf{Personas} Each metric considers a set of personas, $P$, that are provided to the model. Each persona is evaluated under multiple experimental settings, and is expected to yield slightly different results that, in turn, affect performance, outcome, and correct question instability metrics in different ways.

\noindent\textbf{Persona Accuracy} 
The accuracy of a persona $p$ is calculated 
for a set of questions $Q$ under an experimental setting $s$ denoted as,
$\textit{Acc}(p ; s, Q)$. Note that the question set $Q$ could be the entire dataset, or specific subsets, such as domain-specific questions.

\noindent\textbf{Persona Rank} 
Given persona accuracy $\textit{Acc}(p ; s, Q)$, the rank of a persona $p$ relative to other personas is defined as the index of $p$ in a list ranked by accuracy, denoted as $R(p ; s, Q)$.\footnote{In the rest of section~\ref{sec:metric_formulation}, we omit $Q$ in accuracy $\textit{Acc}(p ; s)$ and rank $R(p ; s)$ as $Q$ is implied in computing these metrics.}

\subsection{Performance Instability: $I_A$} The first metric aims to understand how sensitive model accuracy is across all experimental settings (averaged over personas).
Intuitively, we seek to understand that, if we randomly selected two experimental settings from a set of potential settings, what is the expected difference in accuracy?
\begin{align*}
I_A(p) 
  &= \frac{1}{\binom{|S|}{2}}
     \sum_{s_i,\, s_j \in S} 
     \bigl| \operatorname{Acc}(p; s_i) - \operatorname{Acc}(p; s_j) \bigr| \\[6pt]
I_A 
  &= \frac{1}{|P|} 
     \sum_{p \in P} I_A(p)
\end{align*}

The formulation is similar to~\citet{zhuo-etal-2024-prosa-promptsensiscore}, but fitted to a persona-specific context. For a  persona, we calculate the pair-wise accuracy differences between $s_i$ and $s_j$ for all experiment settings, then average them. $\binom{|S|}{2}$ represents the count of experiment settings pairs for a single persona. We aggregate the persona instability to represent the accuracy instability $I_A$ for a given question set, and a model, where $|P|$ is the total number of personas.

\subsection{Outcome Instability $I_O$} This metric measures how variable the conclusion of a persona-driven experiment is, such as identifying the best or worst persona for a task. To do this, we compare persona ranking $R(p ; s)$ across experimental settings. Importantly, we note that for each experimental setting, both minor and substantial differences in persona average accuracies can change their rankings, leading to outcome instability. However, intuitively, we argue that the degree of this instability is different: larger accuracy differences should have a larger outcome instability score. To account for this, we define the outcome instability of each persona as the weighted standard deviation of its rankings, where the weight assigned to each experimental setting corresponds to the standard deviation across all personas' accuracies in that setting. At the model level, overall outcome instability is computed by averaging the persona-level instabilities and scaling by the average of $w_s$ across experimental settings, accounting for models with greater persona parity overall.

\begin{align*}
w(s) 
  &= \operatorname{Std}\!\bigl(\operatorname{Acc}(p_i; s)\bigr),
 \quad  p_i \in P \\[6pt]
I_O(p) 
  &= \operatorname{Std}_{w}\!\bigl(R(p; s_j),\, w(s_j)\bigr),
  \quad  s_j \in S   \\[6pt]
I_O 
  &= \frac{1}{|P|}\sum_{p_i \in P} I_O(p_i) 
     \;\cdot\; 
     \frac{1}{|S|}\sum_{s_j \in S} w(s_j)
\end{align*}

In the above equations, $w(s)$ is calculated with the standard deviation of persona accuracies $\operatorname{Acc}(p_i; s)$ across all personas. The outcome instability of a persona $I_O(p)$ is computed with the weighted standard deviation~\cite{seabold2010statsmodels} of the persona's rank $R(p; s_j)$ using the weight $w(s_j)$ across all experiment settings $s_j$. $I_O$ is the overall model-level outcome instability computed using $I_O(p)$ and $w(s)$.\footnote{We experimented with several alternative formulations of $I_O$; see Appendix~\ref{app:i_o_formulations} for details.}

\subsection{Question Correctness Instability $I_Q$} 
The third metric evaluates the variability of the specific questions that a persona-driven generation answers correctly across experimental settings. Personas associated with a task domain, for example, a biologist in biology, should consistently answer domain-relevant questions correctly; however, the performance may still be sensitive to the experimental settings. Thus, we calculate the percent of questions correct that are shared across all experimental settings given a persona, seeking to answer: How consistent are persona-specific performances across experimental settings?

\begin{align*}
    CQ(p; s) & = \{\, q \in Q \;\mid\; q \text{ is correct using $p$ in $s$} \,\} \\[6pt]
I_Q \;& =\; 1 - \frac{1}{|P|} \sum_{p \in P} 
      \frac{\bigl|\;\bigcap_{s \in S} CQ(p; s)\;\bigr|}
           {\bigl|\;\bigcup_{s \in S} CQ(p; s)\;\bigr|}
\end{align*}

The above equation quantifies the calculation of $I_Q$. Here, $CQ(p; s)$ represents the set of questions correctly answered by persona $p$ under experimental setting $s$. For each persona, we first compute the Jaccard similarity of these sets across all settings, reflecting the consistency of prediction on a question level. This is averaged over personas. To align with the interpretation of the other instability metrics, we subtract this value from 1, to represent the percentage of questions that are not shared (thus greater value indicates higher instability).

%% file: latex/3_figure_tex/demonstration_of_metrics.tex
\begin{figure*}
    \centering
    \includegraphics[width=\textwidth]{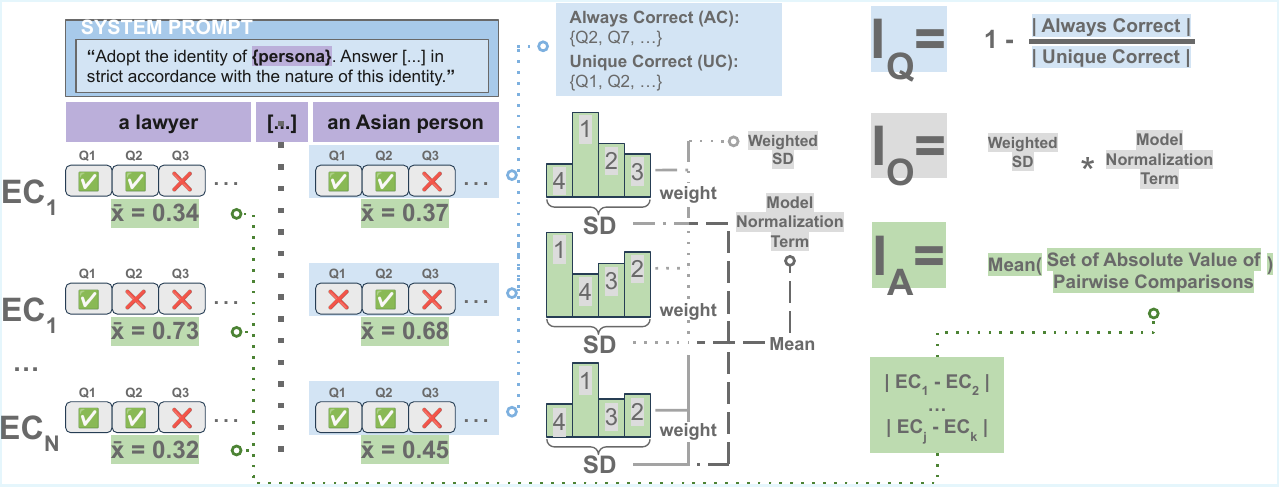}
    \caption{A figure depicting how instability metrics are calculated based on PDGs for our dataset. Across runs in several unique experimental configurations (ECs), we consider mean performance and question-wise correctness. Mean performance is used for standard deviation calculations for computing $I_A$ and used to get inter-persona rankings within one EC for $I_O$. The set of correct questions for a persona across all ECs is used for $I_Q$ calculation. Calculations shown are for one persona, although all personas are integrated as shown in Section~\ref{sec:metric_formulation}. We explore how varying the question set and experiment setting set sizes affects metric calculations in Appendices~\ref{app:sensitivity-question-count} and \ref{app:sensitivity-setting-count}.}
    \label{fig:dataset_volatility}
\end{figure*}

%% file: latex/0_sections/4_experiments.tex
\section{Experiments}
\input{latex/1_tables/model_baselines}
\input{latex/1_tables/persona_table_with_less_example}
\input{latex/1_tables/dataset_table_with_less_example}

\label{sec:experiments}

For experimentation, we evaluate 4 open-source LLMs on an aggregated dataset with questions from different domains under a set of experimental settings for instability calculations and analyses. 
We introduce our persona, dataset and models below. Section~\ref{sec:experimetal_setting_parameters} explains the hyperparameters used to derive all experimental settings. 

\noindent\textbf{Personas}
\label{sec:personas}
We consider 41 personas, split across 9 distinct groups centering on aspects of identity (such as race) and expertise (such as occupation) for comparison. As we seek to compare stability of PDGs across experiment settings similar to those in past literature, these personas are also drawn from previous work, with additional personas added in order to ensure inclusivity and better representation per category~\cite{zheng-jurgens-etal-2024-helpful, gupta2024personabias, wan-etal-2023-personalized-jieyu-stochastic-parrots}. The personas and categorization can be seen in Table~\ref{tab:full-persona-table}. We use both occupation-related personas and demographic-related personas due to their usage in past PDG work, and to improve the generalizability of our findings to future PDG studies. This is further discussed in Section~\ref{sec:ethical_considerations}, with desiderata for stability based on the persona used discussed in Appendix~\ref{app:desiderata-for-stability}.

\noindent\textbf{Dataset}
\label{sec:dataset}
To evaluate and understand PDG instability in LLMs, we create a composite evaluation dataset, $\mathcal{D}$, balanced across various task types and domains for our experiments. We pull from MMLU~\cite{hendrycks2020measuring-mmlu}, Social IQa~\cite{sap-etal-2019-social-socialiqa-siqa}, and NormAd-Eti~\cite{rao-etal-2025-normad}. We group questions under 10 different categories, composed of questions with specific sublabels from MMLU, Social IQa, and NormAd-Eti. The categories and example labels are in Table~\ref{tab:dataset-table}. Further details for dataset and persona are in Appendix~\ref{app:dataset_construction}.

\noindent\textbf{Models}
\label{sec:models}
We evaluate five open-source models: Llama3.2-1B-Instruct and Llama3.1-8B-Instruct~\cite{grattafiori2024llama}, Qwen2.5-1.5B-Instruct, Qwen2.5-7B-Instruct, and Qwen2.5-14B-Instruct~\cite{yang2025qwen3}. These were chosen due to allow comparison across model size and family.

\subsection{Experiment Setting Hyperparameters}

\label{sec:experimetal_setting_parameters}
We consider three experiment setting hyperparameters, drawing from noted differences from prior work~\cite{gupta2024personabias, zheng-jurgens-etal-2024-helpful, de2025principled}. We evaluate our models across all combinations of these hyperparameters, resulting in a total of 48 experiment settings.

We seek to understand instability due to \textit{hyperparameter} variations in this persona context. We additionally analyze instability due to \textit{persona} variations in Appendix~\ref{subsec:persona-as-probe-for-bias}.

\noindent\textbf{Persona Prompt Format} (PPF) is the format of the system prompt used to give an LLM a persona. We draw upon variants from past work that varied in length and abstractness, utilizing four PPF settings~\cite{gupta2024personabias, zheng-jurgens-etal-2024-helpful}.

\noindent\textbf{Task Prompt Format} (TPF) is the format of the prompt used to give an LLM a question. Similar to PPF, we draw upon variants from past work~\cite{gupta2024personabias, zheng-jurgens-etal-2024-helpful, wei2022chain}
that varied in length and expressiveness, considering four TPF settings. Both PPF and TPF examples are shown in Table~\ref{tab:prompt-table} in the Appendix.

\noindent\textbf{Temperature}, or model temperature ($\tau$), is considered. We utilize three different settings -- 0, 0.5, and 1.0 -- that have been observed in previous work~\cite{gupta2024personabias, zheng-jurgens-etal-2024-helpful}.

%% file: latex/1_tables/model_baselines.tex
\begin{table*}[]
\centering
\small
\begin{tabular}{@{}ccllll@{}}
\toprule
\textbf{Model}                             & \textbf{$\uparrow$ Mean Acc (\%)} & \multicolumn{1}{l}{\textbf{$\downarrow$ STD Acc (\%)}} & \multicolumn{1}{l}{\textbf{$\downarrow$ $I_A$}} & \multicolumn{1}{l}{\textbf{$\downarrow$ $I_O$}} & \multicolumn{1}{l}{\textbf{$\downarrow$ $I_Q$}} \\ \midrule
\multicolumn{1}{c|}{Llama-3.2-1B-Instruct} & \multicolumn{1}{c|}{35.101}       & \multicolumn{1}{c|}{3.589}                             & \multicolumn{1}{c|}{3.697}                      & \multicolumn{1}{c|}{18.791}                     & 99.960                                          \\ \midrule
\multicolumn{1}{c|}{Llama-3.1-8B-Instruct} & \multicolumn{1}{c|}{63.287}       & \multicolumn{1}{c|}{2.647}                             & \multicolumn{1}{c|}{2.999}                      & \multicolumn{1}{c|}{5.784}                      & 78.293                                          \\ \midrule
\multicolumn{1}{c|}{Qwen2.5-1.5B-Instruct} & \multicolumn{1}{c|}{43.714}       & \multicolumn{1}{c|}{5.826}                             & \multicolumn{1}{c|}{6.586}                      & \multicolumn{1}{c|}{14.305}                     & 98.576                                          \\ \midrule
\multicolumn{1}{c|}{Qwen2.5-7B-Instruct}   & \multicolumn{1}{c|}{67.771}       & \multicolumn{1}{c|}{0.963}                             & \multicolumn{1}{c|}{0.972}                      & \multicolumn{1}{c|}{4.539}                      & 57.688                                          \\ \midrule
\multicolumn{1}{c|}{Qwen2.5-14B-Instruct}  & \multicolumn{1}{c|}{73.668}       & \multicolumn{1}{c|}{1.433}                             & \multicolumn{1}{c|}{1.623}                      & \multicolumn{1}{c|}{4.812}                      & 52.828                                         
\end{tabular}
\caption{A table depicting the baseline overall instability results for each model, reporting mean and standard deviation of accuracy across all runs, and our three instability metrics. It can be noted that larger model size leads to less instability across all dimensions, and differences in instability between model families.}
\label{tab:model_baselines}
\end{table*}

%% file: latex/1_tables/persona_table_with_less_example.tex
\begin{table}[!t]
\centering
\tiny
\begin{tabular}{@{}c|c|l@{}}
\toprule
\textbf{Persona Category} & \textbf{\begin{tabular}[c]{@{}c@{}}Total \#\\ Personas\end{tabular}} & \multicolumn{1}{c}{\textbf{Example Personas}} \\ \midrule
Gender                    & 5                                                                    & a man, a non-binary person                    \\ \midrule
Sexuality                 & 3                                                                    & a straight person, a gay person,              \\ \midrule
Race/Ethnicity            & 6                                                                    & an African person                             \\ \midrule
Age                       & 3                                                                    & an adult, a young person                      \\ \midrule
Disabilities              & 2                                                                    & a disabled person                             \\ \midrule
Religious Beliefs         & 4                                                                    & a religious person                            \\ \midrule
Political Beliefs         & 4                                                                    & a Democrat, a Republican                      \\ \midrule
Occupational Roles        & 12                                                                   & a biologist, a lawyer                         \\ \midrule
Baselines                 & 2                                                                    & a human, NO PERSONA                           \\ \bottomrule
\end{tabular}
\caption{Example personas used in experimentation.}
\label{tab:persona-table}
\end{table}

%% file: latex/1_tables/dataset_table_with_less_example.tex
\begin{table}[!t]
\centering
\tiny
\begin{tabular}{@{}ccl@{}}
\toprule
\textbf{Dataset Category}                                   & \textbf{\begin{tabular}[c]{@{}c@{}}Total \#\\ Questions\end{tabular}} &  \multicolumn{1}{c}{\textbf{Example Sub-Labels}}                                         \\ \midrule
\multicolumn{1}{c|}{History}                                & \multicolumn{1}{c|}{500}                                                                                              & \begin{tabular}[c]{@{}l@{}}'prehistory'\end{tabular}     \\ \midrule
\multicolumn{1}{c|}{Politics and Law}                       & \multicolumn{1}{c|}{500}                                                                                              & \begin{tabular}[c]{@{}l@{}}'us\_foreign\_policy'\end{tabular}    \\ \midrule
\multicolumn{1}{c|}{Social Science}                         & \multicolumn{1}{c|}{497}                                                                                             & 'sociology'                                                               \\ \midrule
\multicolumn{1}{c|}{Business/Accounting and Economics}      & \multicolumn{1}{c|}{496}                                                                                              & 'econometrics'                                                      \\ \midrule
\multicolumn{1}{c|}{EECS}                                   & \multicolumn{1}{c|}{500}                                                                                               & \begin{tabular}[c]{@{}l@{}}'machine\_learning'\end{tabular}     \\ \midrule
\multicolumn{1}{c|}{Math}                                   & \multicolumn{1}{c|}{498}                                                                                             & \begin{tabular}[c]{@{}l@{}}'formal\_logic'\end{tabular}       \\ \midrule
\multicolumn{1}{c|}{Natural Science}                        & \multicolumn{1}{c|}{496}                                                                                            & \begin{tabular}[c]{@{}l@{}}'high\_school\_chemistry'\end{tabular} \\ \midrule
\multicolumn{1}{c|}{Human Health and Medicine}              & \multicolumn{1}{c|}{504}                                                                                               & 'college\_medicine'                                                           \\ \midrule
\multicolumn{1}{c|}{Cultural Awareness and Understanding}   & \multicolumn{1}{c|}{525}                                                                                              & 'normad\_iraq'                                                       \\ \midrule
\multicolumn{1}{c|}{Commonsense Reasoning/Social Awareness} & \multicolumn{1}{c|}{497}                                                                                              & 'motivations'                                                                \\ \bottomrule
\end{tabular}
\caption{Dataset categories used in experimentation.}
\label{tab:dataset-table}
\end{table}


%% file: latex/0_sections/5_results_analysis.tex
\section{Results and Analysis}
\label{sec:results_analysis}

\input{latex/0_sections/5_analysis_subsections/1_analysis_baselines}
\input{latex/0_sections/5_analysis_subsections/2_analysis_parameter_wise}
\input{latex/0_sections/5_analysis_subsections/3_analysis_parameter_interactions}
\input{latex/0_sections/5_analysis_subsections/4_analysis_by_category}

\input{latex/0_sections/5_analysis_subsections/5_analysis_accuracy}
\input{latex/0_sections/5_analysis_subsections/6_analysis_different_experimental_outcomes}

%% file: latex/0_sections/5_analysis_subsections/1_analysis_baselines.tex
\subsection{Baselines}
\input{latex/3_figure_tex/stratified_parameter_heatmap_parameter_volatility_both_models}

We perform baseline experiments for all 41 personas and 48 experimental settings per model. Both model performance and instability are shown in Table~\ref{tab:model_baselines}. Note that for all metrics, a perfectly stable model would yield a score of 0, indicating consistent performance across different hyperparameter settings. In practice, hyperparameters inevitably cause fluctuations, so non-zero values are expected.

As shown in Table~\ref{tab:model_baselines}, all evaluated models' instability measurements are well above 0, with particularly high $I_Q$ scores indicating the limited overlap in correctly answered questions between experiment settings, like less than 1\% for the Llama 1B model. Secondly, higher mean accuracy largely corresponds to lower accuracy instability: Llama 1B and Qwen2.5-1.5B show relatively low accuracy and a high $I_A$, in contrast to their larger counterparts with a high accuracy and a lower $I_A$. Outcome instability $I_O$ shows a clear trend with both model performance and size, with larger models demonstrating greater stability across all metrics\footnote{The only exception to this trend is Qwen2.5-14B. We discuss our rationale for this pattern in Appendix~\ref{app:large_qwen_results}.}. We refer to these results as the baseline results. To better compare across models and model families, all subsequent results are reported relative to these baselines. Specifically, we compute the instability difference as (baseline – new), where a positive value indicates lower instability in the new experiments and an improvement over the baseline.

%% file: latex/3_figure_tex/stratified_parameter_heatmap_parameter_volatility_both_models.tex
\begin{figure*}
    \centering
    \includegraphics[width=\textwidth]{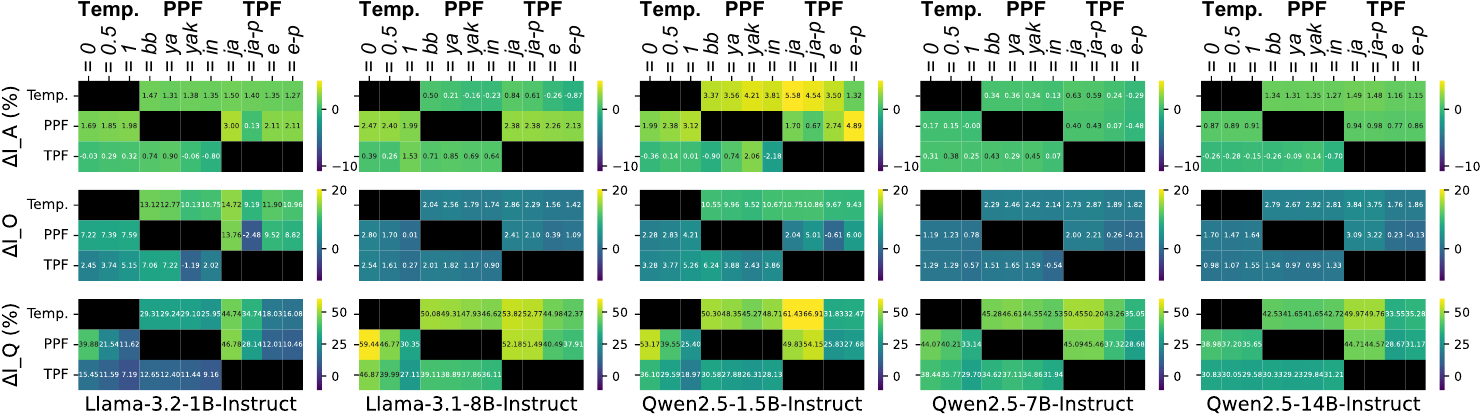}
    \caption{A figure depicting the delta hyperparameter instability of temperature (Temp.), persona prompt format (PPF), and task prompt format (TPF) compared to the overall baseline, stratified by experimental settings with specific settings for each hyperparameter. Here, a greater value (brighter) means less instability compared to the baseline. A key for the column labels is in Appendix~\ref{app:heatmap-label-keys}.}
    \label{fig:stratified_volatility}
\end{figure*}

%% file: latex/0_sections/5_analysis_subsections/2_analysis_parameter_wise.tex
\subsection{Instability From Each Hyperparameter}
\label{subsec:single parameter}
\input{latex/3_figure_tex/grouped_bar_plot_parameter_volatility_no_spread}
\input{latex/3_figure_tex/heatmap_by_both_categories}

We aim to investigate the role of single hyperparameter variations ($\tau$, PPF, and TPF) in instability. Here, the experimental settings set $S$ is built by varying one hyperparameter at a time (e.g., $\tau$ at 0.0, 0.5, 1.0) while holding others fixed. For each hyperparameter, results are averaged across all fixed settings. Figure~\ref{fig:parameter_volatility} shows the difference between the instability scores to baseline scores for all models.

We see that compared to the baseline, nearly all hyperparameter-specific instability scores are lower, as indicated by positive values. This shows that varying a single hyperparameter makes the model more stable than varying all hyperparameters simultaneously.
Secondly, we observe that all hyperparameters ($\tau$, PPF, and TPF) produce similar trends across models, with $\tau$ typically yielding the highest values (greatest improvements) and TPF the lowest. Since differences in instability scores reflect reductions in instability, smaller decreases indicate greater instability in model predictions. Thus, since TPF shows the smallest differences, we argue that this makes it the most influential hyperparameter for overall instability.
We also note that different instability metrics show varying trends: some, like Llama 8B’s $I_Q$ between PPF and TPF, remain relatively consistent, while others, such as $I_A$, have steep drops. This shows the importance of instability metrics from different perspectives.


%% file: latex/3_figure_tex/grouped_bar_plot_parameter_volatility_no_spread.tex
\begin{figure}[!h]
    \centering
    \includegraphics[width=\columnwidth]{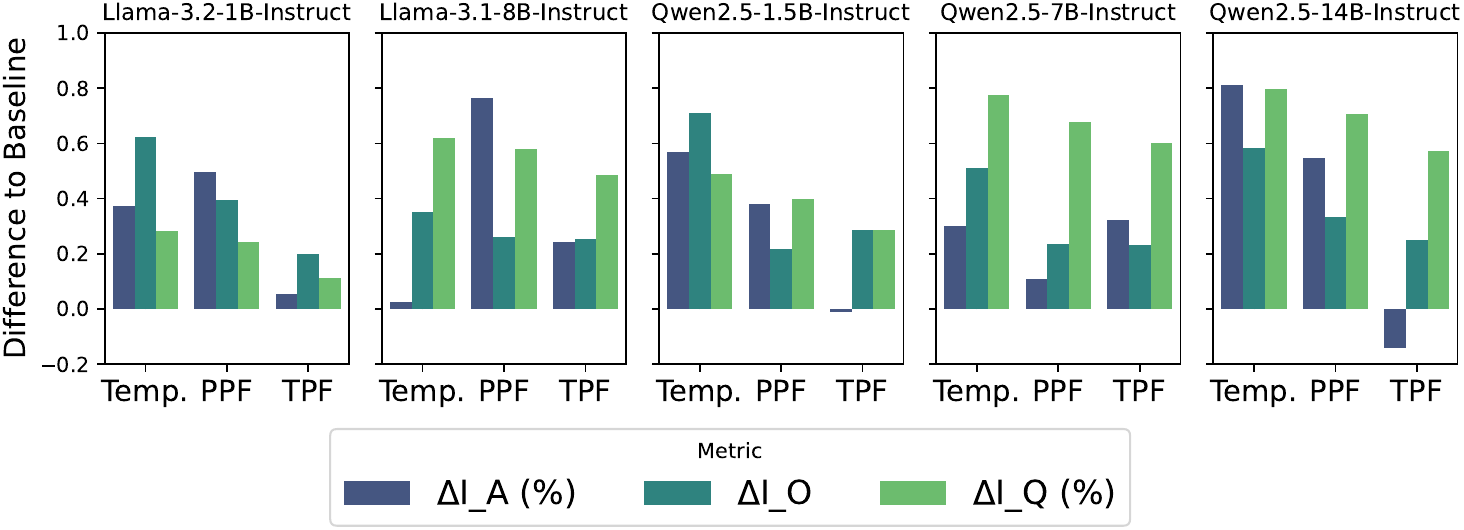}
    \caption{Depicting the delta hyperparameter instability of Temp., PPF, and TPF compared to the baseline. Here, a greater value means less instability compared to the baseline. We report results for each hyperparameter divided by those of the baseline to improve readability.
    }    
    \label{fig:parameter_volatility}
\end{figure}

%% file: latex/3_figure_tex/heatmap_by_both_categories.tex
\begin{figure*}
    \centering
    \includegraphics[width=\textwidth]{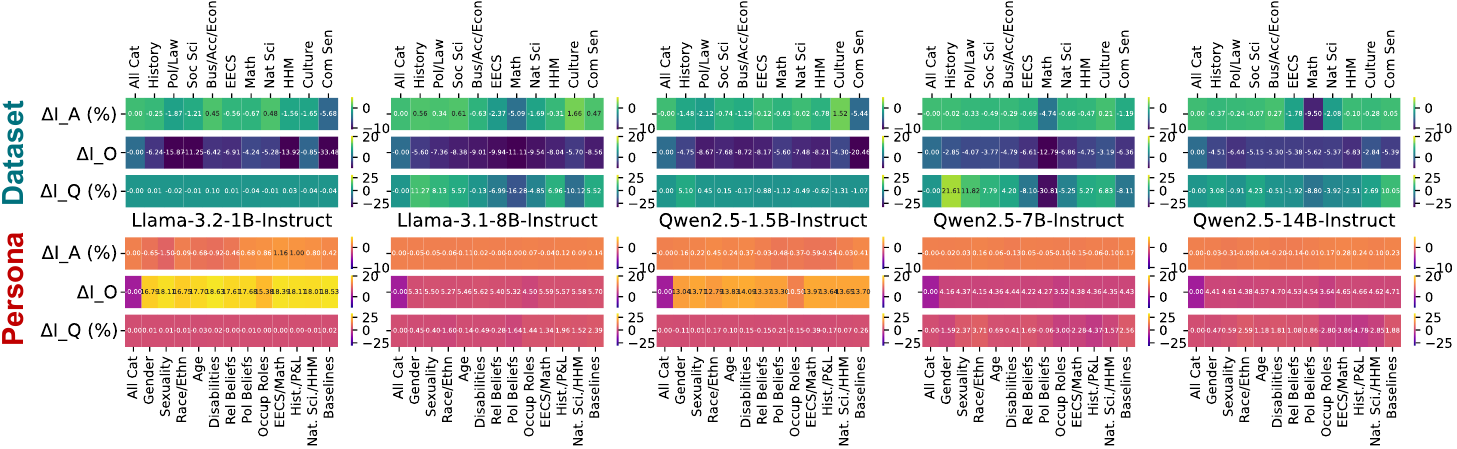}
    \caption{Depicting the delta hyperparameter instability of Temp., PPF, and TPF compared to the baseline. "Dataset" and "Persona" are stratified by the question and persona categories, respectively. A label key is in Appendix~\ref{app:heatmap-label-keys}.}
    \label{fig:dataset_and_persona_volatility}
\end{figure*}


%% file: latex/0_sections/5_analysis_subsections/3_analysis_parameter_interactions.tex
\paragraph{Which Value Causes the Highest Instability for Each Hyperparameter?}
\label{subsec:parameter_interactions}

Here, we seek to identify specific "unstable hyperparameter settings" for each hyperparameter that may lead to increased instability. We perform the same procedure as for single-hyperparameter instability, except we stratify results by experimental settings with specific hyperparameter settings for each hyperparameter (e.g. finding instability when varying TPF in settings with $\tau = 0$). 
Figure~\ref{fig:stratified_volatility} shows the results.

From this, we identify key "unstable" and "stable" settings for each hyperparameter. We find that, for $\tau$ (first three columns), a $\tau=0$ is the most stable (lighter color), exhibiting higher stability across nearly all models and metrics -- a $\tau=1$ exhibits the opposite trend (darker color), especially as seen in $I_A$. This aligns with the intuition that a higher temperature generate more diverse predictions, i.e different answers across runs. For PPF, we see the more verbose or abstract PPFs have generally lower stability, with "barebones" being the most stable for nearly all models and metrics, and "youarknowledge" or "identitynature" being the least stable. For TPF, we observe a similar trend, with the TPFs requiring more detail -- the "explain" variations -- having the least stability across nearly all models and metrics. We note that an unstable TPF format causes greater instability than an unstable temperature value, mirroring results in Section~\ref{subsec:single parameter}.

%% file: latex/0_sections/5_analysis_subsections/4_analysis_by_category.tex
\subsection{Instability Across Categorization} 
\label{subsec:category-analysis}

We investigate how instability varies based on question and persona categories. We follow the same procedure as for single-hyperparameter variations, except further stratifying with dataset categories or persona categories. The results are in Figure~\ref{fig:dataset_and_persona_volatility}, with the top sub-figure showing dataset trends and the bottom sub-figure showing  persona trends.

\noindent\textbf{Dataset} In the top figure, we observe metric- and domain-specific trends.
For $I_A$, we see considerable variation across categories, with many dataset categories having negative scores (higher instability compared to baseline) and only a few having positive scores (lower instability compared to baseline), indicating that model performance fluctuates considerably between domains. We can additionally observe a relationship between $I_A$ and $I_Q$, where in larger models they exhibit nearly identical patterns. In smaller models, $I_Q$ is largely unchanged, possibly due to the initial large variation of question overlaps overall and domain-specific experiments exhibiting similar instability with questions. $I_O$ is completely negative. We hypothesis this is due to the smaller number of domain-specific questions compared to the entire dataset, so differences in question correctness lead to greater total performance differences and variations in ranking.  Thus, the domain-specific $I_O$ will be lower.

Across dataset domains, we observe that math-related and commonsense questions lead to the highest instability (darker color), whereas history and cultural awareness/understanding lead to the lowest instability (lighter color) across nearly all metrics and models. We hypothesize that this can be attributed to disparities in training data relative to these domains: commonsense knowledge is often underrepresented due to reporting bias~\cite{gordon2013reporting}, while mathematical content constitutes only a small fraction of most models’ training corpora~\cite{elazar2024whats}.\footnote{We analyzed the format and length of the questions from different categories, but see no relationship between instability and questions of similar style, length, or answer choice count.}

\noindent\textbf{Persona} In the bottom figure in Figure~\ref{fig:dataset_and_persona_volatility}, we see little relation between persona category (e.g., what personas are used in experimentation) and instability, with relatively equivalent results across all metrics, models, and persona categories. This shows that the instability of PDGs in MCQA has less to do with the \textit{specific} persona used, but rather the hyperparameters used in the experiment settings.\footnote{We note differences in $I_O$. We attribute this to the differences in persona category size (with Gender, Race/Ethnicity, and Occupational Roles being the largest). A greater number of personas confers greater maximal ranking shifts, influencing $I_O$. Similar-sized persona groups have similar $I_O$.}

This finding is bolstered by further analysis with sub-categories of occupational personas. We sub-categorize 9 occupational personas into groups of three, with each group having occupations with expertise closely associated with two categories in $\mathcal{D}$ -- EECS/Math, History/Politics and Law, and Natural Science/Human Health and Medicine. This categorization is described in Appendix~\ref{app:persona-details}. We note little differences in instability when comparing these sub-categories to other persona categories, supporting our previous claims. Further details on this analysis and extensions are in Appendix~\ref{app:sub-category-dataset-category}.


%% file: latex/0_sections/5_analysis_subsections/5_analysis_accuracy.tex
\subsection{Stability, Accuracy, and Results}
\label{subsec:stability-and-accuracy}
\input{latex/1_tables/stability_and_accuracy} 

We compare the relationship between stability, as identified by our metrics, and overall accuracy on this task. Using the low and high stability hyperparameter settings observed in Section~\ref{subsec:parameter_interactions}, we identify four stable and unstable experiment settings: 
\textbf{Stable} -- $\tau$ = 0.0, TPF = \{"justanswer", "justanswer-persona"\}, PPF = \{"barebones", "youare"\};
\textbf{Unstable} -- $\tau$ = 1.0, TPF = \{"justanswer", "justanswer-persona"\}, PPF = \{"barebones", "youare"\}.

We compare the mean accuracy and maximal accuracy across all stable settings versus all unstable settings, averaged across all personas, reporting results in Table~\ref{tab:stability_and_accuracy}. As seen in Table~\ref{tab:stability_and_accuracy}, stable settings largely outperform unstable settings, exhibiting greater mean and maximal accuracy\footnote{The only exception to this trend is Qwen2.5-14B. We discuss our rationale for this pattern in Appendix~\ref{app:large_qwen_results}.}. This cements a relationship between stability and model performance, and emphasizes the importance of considering stability for LLM PDGs.

%% file: latex/1_tables/stability_and_accuracy.tex
\begin{table}[h]
\small
\centering
\begin{tabular}{@{}l|ll|ll@{}}
\toprule
\textbf{}       & \multicolumn{2}{l|}{\textbf{Stable Settings}} & \multicolumn{2}{l}{\textbf{Unstable Settings}} \\ \midrule
\textit{Models} & \textbf{Average}       & \textbf{Max}         & \textbf{Average}        & \textbf{Max}         \\ \midrule
Llama-3.2-1B    & \textbf{0.382}         & \textbf{0.404}       & 0.318                   & 0.331                \\
Llama-3.1-8B    & \textbf{0.634}         & \textbf{0.635}       & 0.610                   & 0.615                \\
Qwen2.5-1.5B    & \textbf{0.499}         & \textbf{0.555}       & 0.366                   & 0.393                \\
Qwen2.5-7B      & \textbf{0.681}         & \textbf{0.682}       & 0.663                   & 0.674                \\
Qwen2.5-14B     & 0.728                  & 0.731                & \textbf{0.745}          & \textbf{0.749}       \\ \bottomrule
\end{tabular}
\caption{Depicting the average and maximal accuracy across stable and unstable settings per model.}
\label{tab:stability_and_accuracy}
\end{table}

%% file: latex/0_sections/5_analysis_subsections/6_analysis_different_experimental_outcomes.tex
\paragraph{Results and Instability}
\input{latex/3_figure_tex/example_different_experiment_outcomes}
We compare results between low and high stability experiment settings, using a similar procedure. We consider the stable setting $\tau$ = 0, PPF = "barebones", and TPF = "justanswer", and unstable setting $\tau$ = 1, PPF = "identitynature", and TPF = "explain-persona".

Between the two settings, \textbf{1)} the best- and worst-performing personas across the entire dataset differ, and \textbf{2)} the distribution of persona performances differs. This cements a completely different results landscape when comparing high and low stability settings as identified through our metrics. This is demonstrated with Llama 1B in Figure~\ref{fig:different_experimental_outcomes}, with all models exhibiting similar patterns, as seen in Figure~\ref{fig:app:all_experiment_outcomes} in the Appendix.\footnote{We additional explore how instability impacts the \textit{predictability} of persona results in Appendix~\ref{subsec:classifier_performance}.}

%% file: latex/3_figure_tex/example_different_experiment_outcomes.tex
\begin{figure}[!h]
    \centering
    \includegraphics[width=\columnwidth]{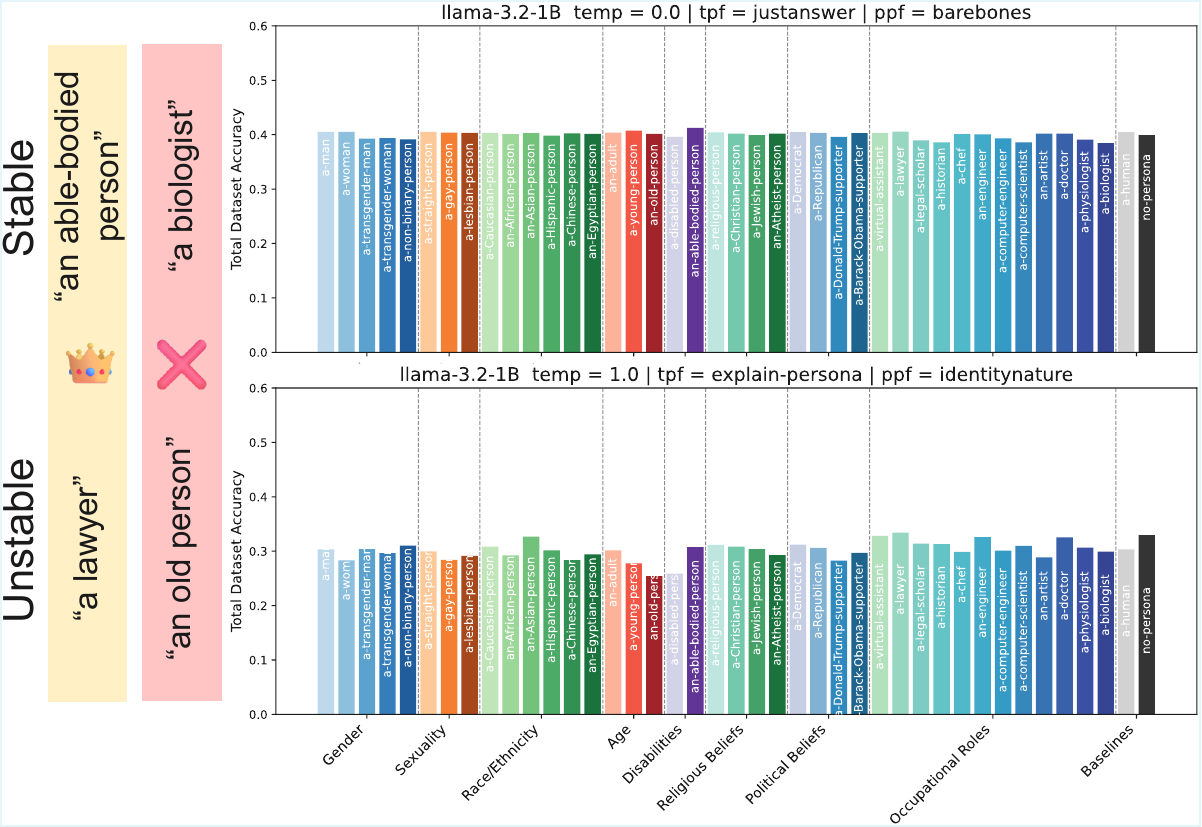}
    \caption{An example of different experimental outcomes for Llama 1B in a stable and unstable setting. The best/worst persona relative to accuracy for each setting are shown in gold and red, respectively. Note: \textcolor{red}{identity/demographic should \textbf{not} impact model performance.} Please see the Limitations and Ethical Considerations.}
    \label{fig:different_experimental_outcomes}
\end{figure}

%% file: latex/0_sections/7_conclusion.tex
\section{Conclusion}


In this paper, we introduce three instability metrics to evaluate LLMs PDGs in MCQA, assessing a model’s sensitivity in performance, outcome, and overlap of correctly answered questions with varying prompts and temperatures.
We use these metrics to understand differences with past work and to inform future PDG experiment designs. We find that PDGs are heavily unstable across 3 distinct dimensions, with similar, although different trends across our three metrics, consistent across all models. We found that models are most sensitive to variations in TPF compared to parameters like PPF or generation temperature. Instability varies across question domain, with math questions exhibiting the highest instability. Additionally, we find a relationship between instability and accuracy, with unstable settings having substantial decreases in accuracy compared to stable settings. Importantly, these unstable hyperparameters can lead to entirely different conclusions in persona-based analyses. We hope that our work supports future PDG experiments and analysis of their instability.




%% file: latex/0_sections/limitations_considerations_acknowledgments.tex
\section*{Limitations}
\label{sec:limitations}

\noindent\textbf{Instability Metric Purpose} We note that these metrics capture instability \textit{in} a persona setting, rather than instability solely \textit{due to} the inclusion of persona, with our metrics capturing instability due to variations in persona, prompt format, etc. The potential for conflating factors -- such as different model temperatures, prompt formats, task content (which can relate to task difficulty), and personas used -- is the subject of analysis within this paper. We discuss how the analysis of these factors impacting instability can be utilized for future studies or usages of PDGs in MCQA, as we identify when these uses may be most unstable. We additionally discuss how a re-formulation of these metrics to consider persona as a parameter can identify new facets of instability (such as instability due to varying \textit{persona}, rather than other hyperparameters) and bias in models in Appendix~\ref{subsec:persona-as-probe-for-bias}.

\noindent\textbf{Persona} We consider both occupational personas, as well as identity-related personas, in order to reflect the distribution of personas used in past work. We acknowledge that the set of personas used is not fully representative of personas used in PDGs in both casual use and past research, and may not capture the full diversity present within personas of each category. We treat our personas used as a broad approximation of the various identities and occupations used with PDGs, drawing heavily from those used in past work, and encourage future work that may be able to expand the set of personas for other investigations. We include details of how we believe stability should manifest across occupation-related and identity-related personas to ensure unbiased performance, user safety, and adequate representation in Appendix~\ref{app:desiderata-for-stability}.

\noindent\textbf{Tasks and Domain} We acknowledge that the tasks and domain of tasks chosen may not fully represent the broad range of tasks that PDGs are used in, in both research and non-research applications. While we focus on multiple-choice answering as our means of assessing PDGs in non-text-heavy settings, we encourage future work that evaluates other non-text-heavy paradigms utilizing PDGs, such as smaller annotation tasks.

\noindent\textbf{Experimental Settings} We acknowledge that the experimental settings and hyperparameters that we investigate do not represent the full list of potential settings or hyperparameters that people may use for their own PDGs. We base our choices off of differences noted in prior work and resource constraints, but surely other variations exists that can have their own specific relationships with instability.

\noindent\textbf{Models} While we chose our selection of models such that we could identify potential family and size-related differences in instability, we do not fully encompass the diversity of model families and sizes accessible today. We encourage future work evaluating model instability in non-text-heavy PDGs across a wider variety of model sizes (such as larger 70B models), or different model architectures.

\section*{Ethical Considerations}
\label{sec:ethical_considerations}

\noindent\textbf{Inclusion of Identity-Related Personas}
As our study relates to aspects of PDG experiments, such as experiment setting, we consider it valuable to evaluate across a diverse set of personas to improve the generalizability of our results. Because of this, we employ not only occupation or expertise-related personas, but also demographic-related personas, as seen in past PDG literature. In this, while we consider accuracy as an important factor of stability for PDGs, \textbf{we make no claim that certain identities should perform better on an MCQA task than others}. We consider identity-related personas impacting accuracy on our MCQA dataset, like "a Hispanic person" impacting performance on math questions, to be undesirable, stereotyped, and biased model behavior. However, as these biased behaviors have been the subject of past studies like~\citet{gupta2024personabias}, we consider it important to evaluate how stable these behaviors may be to inform future studies on this important issue, motivating our inclusion of identity-related personas in our study. 

Additionally, we see how instability can lead to patterns reminiscent of bias, as seen in Figure~\ref{fig:different_experimental_outcomes}. In a scenario where demographic-related personas do not lead to biased outcomes, it is necessary that the lack of bias is generalizable across experiment settings, or stable, for considerations like user safety (while still maintaining variation reflective of individuals -- we discuss this in Appendix~\ref{app:desiderata-for-stability}). For scenarios where demographic personas \textit{do} lead to biased outcomes, stability across experimental settings is still valuable. With instability (as we find), we cannot rely on inter-persona performance differences to suggest a model is biased in a specific way, as small variations in experiment setting impact those resulting conclusions, as seen in Figure~\ref{fig:different_experimental_outcomes}. This may point towards not necessarily human-interpretable biases causing perceived biased performance as in~\citet{gupta2024personabias} (potentially motivating investigations into demographic-related toxicity in training data), but some other mechanism that only manifests in occasionally interpretable ways that mirror biases in our world.

Finally, our stability metrics have direct applications to identifying and understanding these patterns of harmful bias in LLM performance. We are able to identify (to our knowledge) previously unknown facets of biased model performance relative to demographics, being question correctness consistency, as shown in Appendix~\ref{subsec:persona-as-probe-for-bias}. These points justify our usage of both occupational personas and demographic-related personas in our study.

\noindent\textbf{Representation} While we attempt to make our persona selections as inclusive as possible, we recognize that we are unable to capture the complete diversity present within people with identities associated with each persona category. Additionally, people's identities are far more pluralistic, with discrete labels likely not capturing someone's full sense of self -- what we use as persona roles are approximations.

\section*{Acknowledgments}

We would like to thank the following entities for thoughtful discussion, feedback, and support throughout the duration of this work:
\begin{itemize}
    \item This research was supported in part by the University of Pittsburgh Center for Research Computing and Data, RRID:SCR\_022735, through the resources provided. Specifically, this work used the HTC cluster, which is supported by NIH award number S10OD028483.
    \item The PittNLP community, for thoughtful discussion and feedback during paper writing.
    \item Alex Maldonado, for feedback regarding figure design and analyses.
\end{itemize}

\noindent\textbf{Generative AI Use Statement}
We use Generative AI (GenAI) tools, specifically ChatGPT, for minor rephrasing, word choice, and figure editing suggestions. GenAI is not used in any way beyond that, such as for idea generation, analyses, claims, section writing, etc. No text is directly copied from any GenAI service, with everything only being used as a reference that is reviewed by the authors to ensure integrity and correctness.

%% file: latex/0_sections/8_appendix.tex
\appendix

\section{Prompts}
\label{app:prompts}

\subsection{TPF and PPF}
\input{latex/1_tables/prompt_table}
The full prompts used for the task prompt format (TPF) and persona prompt format (PPF) hyperparameter can be seen in Table~\ref{tab:prompt-table}.

\section{LLM Parameter Settings and Usage in Experimentation}
We fix the model parameters across models, only varying the temperature hyperparameter across experiments. Otherwise, we maintain the following settings across all models evaluated:
\begin{itemize}
    \item do\_sample: \textit{True}
    \item top\_k: \textit{50}
    \item top\_p: \textit{1.0}
    \item max\_new\_tokens (from HF pipeline): \textit{1000}
\end{itemize}

For experimentation, we evaluate each LLM on an equal amount of experimental settings and personas, leading to 1,968 full evaluations of $\mathcal{D}$ per model. Specifically, for one combination of persona and experiment setting hyperparameters (a setting for each of $\tau$, PPF, and TPF):
\begin{itemize}
    \item We evaluate models on every question in $\mathcal{D}$.
    \item For each question, we provide a system prompt containing our persona prompt, in the specified format of the PPF setting. We substitute in the specific persona for a set of runs.
    \item With this system prompt, each question is then proposed to the model, formatted in accordance to the TPF setting. The question content, including the formatted answer choices, are substituted in within this specified format.
    \item Thus, for each question in $\mathcal{D}$, we provide a formatted system prompt (our persona prompt) and a formatted user prompt (our task prompt).
\end{itemize}

\section{Licenses and Intended Use}

We utilized the Llama-3.1-8B-Instruct and Llama-3.2-1B-Instruct models during evaluation, complying with Meta's LLAMA 3.1 Community License Agreement and LLAMA 3.2 Community License Agreement.

We utilized the Qwen2.5-1.5B-Instruct and Qwen2.5-7B-Instruct models during evaluation, complying with the Apache 2.0 license that these models were released under.

We utilized the OpenAI batch API for prompting GPT-4o-mini for relational knowledge article generation, complying with OpenAI's terms of use.

We additionally made use of previously created Python packages for processing and analysis:
\begin{itemize}
    \item statsmodels: \cite{seabold2010statsmodels}
    \item SciPy: \cite{2020SciPy-NMeth}
\end{itemize}

\section{Dataset and Personas}
\subsection{Dataset Construction}
\label{app:dataset_construction}
To evaluate and understand PDG instability in LLMs, we create a composite evaluation dataset, $\mathcal{D}$, balanced across various task types and domains for our experiments. This topical diversity in questions allows a balanced overall view of model instability, as well as investigation into the relationship between question domain and volatility. We pull from MMLU~\cite{hendrycks2020measuring-mmlu}, Social IQa~\cite{sap-etal-2019-social-socialiqa-siqa}, and NormAd-Eti~\cite{rao-etal-2025-normad}. We group questions under 10 different categories, composed of questions with specific sublabels from MMLU, Social IQa, and NormAd-Eti. For each category, we sample equally across all question label groups, considering rounding -- with this, we get \textasciitilde500 questions per broader category. The categories and questions per category can be seen in Table~\ref{tab:full-dataset-table}.

We select labels per category by referencing organization from prior work~\cite{gupta2024personabias, zheng-jurgens-etal-2024-helpful}. We ensure that each category has a minimum of three question labels associated with it, ensuring topical diversity within a category, although still pertaining to the broader domain. For categories other than Cultural Awareness and Understanding and Commonsense Reasoning/Social Awareness, we take question labels as questions from specific constituent datasets of MMLU, as shown in Table~\ref{tab:full-dataset-table}.

For Cultural Awareness and Understanding, we take directly from NormAd-Eti~\cite{rao-etal-2025-normad}, using the country context paradigm within the dataset. Each country label in the dataset (e.g. "mexico") for a question are used as the question labels for $\mathcal{D}$.

For Commonsense Reasoning/Social Awareness, we follow the categorization of questions from Social IQa based on ATOMIC~\cite{sap-etal-2019-social-socialiqa-siqa}. Using the template provided in this work, we perform rounds of automatic labeling of Social IQa questions with respect to these labels. Following five rounds of labeling and integration of additional templates that match one of the Social IQa question types, we group the remaining questions under "misc."
\input{latex/1_tables/full_dataset_table}

\subsection{Personas}
\label{app:persona-details}
We consider 41 personas for our experimentation, split across 9 distinct groups centering on aspects of identity (such as race) and expertise (such as occupation) for comparison. These personas are drawn upon previous work, with additional personas added in order to ensure inclusivity and better representation per category~\cite{zheng-jurgens-etal-2024-helpful, gupta2024personabias, wan-etal-2023-personalized-jieyu-stochastic-parrots}. Specifically, we take a sample of occupational personas, chosen to consider broad expertise~\cite{zheng-jurgens-etal-2024-helpful}, and an assortment of identity-related personas from other prior work~\cite{gupta2024personabias, wan-etal-2023-personalized-jieyu-stochastic-parrots}, along with additional personas we add ourselves. The complete personas and categorization can be seen in Table~\ref{tab:full-persona-table}.

We additionally consider sub-categories of occupational personas for expertise-wise analysis. With this, we consider three groups of three personas, with each group related to two categories in $\mathcal{D}$. Specifically --  an engineer, a computer engineer, and a computer scientist $\rightarrow$ EECS/Math; a lawyer, a legal scholar, and a historian $\rightarrow$  History/Politics and Law; and a doctor, a physiologist, and a biologist $\rightarrow$ Natural Science/Human Health and Medicine. We also consider a miscellaneous (Misc.) category for comparison, containing: an artist, a chef, and a virtual assistant.
\input{latex/1_tables/full_persona_table}

\subsubsection{Stability Desiderata Depending on Persona}
\label{app:desiderata-for-stability}
We acknowledge that stability desiderata may, and should, vary based on the type of persona used for PDGs in MCQA. The following discussion closely relates to our analysis of stability by persona category, as seen in Section~\ref{subsec:category-analysis}.

For occupational personas or task-aligned personas (such as "a biologist" for biology questions), we believe there should be a great initial stability within these personas' performances. Task-aligned personas should \textit{always} confer a task advantage compared to non-task-aligned personas (similar to what is discussed in~\citet{de2025principled}), with this advantage being consistent across similar environments. Therefore, not only should overall instability be low for task-aligned personas (an analysis shown in Figure~\ref{fig:dataset_and_persona_volatility}), but instability within these personas should not vary across similar experiment settings (an analysis shown in Figure~\ref{fig:app:full_category_analysis}.

Unfortunately, we observe that considering only task-aligned personas still leads to considerable instability, as shown in Figure~\ref{fig:dataset_and_persona_volatility}, even specifically in the task domain, as shown in Figure~\ref{fig:app:sub_and_data_category}. We also see how varying different specific experiment setting hyperparameters can lead to differing patterns of instability in task-aligned personas for the smaller models we evaluate, as shown in Figure~\ref{fig:app:full_category_analysis}.

For identity-related personas, we believe there should be instability or variation within these personas' performances. Variation here is beneficial, as it implies the model does not have a fixed (potentially stereotyped) representation of how a member of an identity group would perform on an MCQA task. This supports individuality and a pluralistic perspective of individuals within an identity group. However, similar to occupational or task-aligned personas, this variation should be consistent across similar environments, with a similar level of variation even with small perturbations in the experiment setting. Additionally, this variation should be consistent across the various identity-related persona categories -- a bias in this regard suggests harmful, differing perspectives of the degree of "individuality" within a specific identity group, which we consider undesirable.

We observe that considering only identity-related personas still displays instability/variation in model performance, as shown in Figure~\ref{fig:dataset_and_persona_volatility}. This may suggest support for individualistic performance that may resemble human variations within these identity groups.\footnote{However, we maintain our perspective that giving a model an identity-related persona should not meaningfully impact performance on irrelevant tasks, like math MCQA, in the first place, to mitigate harm to users.}\footnote{We do not evaluate whether this variation is similar to that of a human population, with this being out of the scope of our study.} However, we do note differences, although minor, in the stability between identity-related persona categories, especially in the smaller models we evaluate. Additionally, as shown in Figure~\ref{fig:app:full_category_analysis}, we see how varying different specific experiment setting hyperparameters can lead to differing patterns of instability in identity-related personas, with these patterns additionally varying \textit{between} the identity-related persona categories. We consider this undesirable and potentially harmful to users.


\section{Additional Analyses}

\input{latex/0_sections/5_analysis_subsections/7_analysis_persona_as_parameter}
\input{latex/0_sections/5_analysis_subsections/8_sub_occupation_category_dataset_category_analysis}
\input{latex/0_sections/5_analysis_subsections/9_analysis_classifier}

\input{latex/0_sections/5_analysis_subsections/10_analysis_correlation}
\input{latex/0_sections/5_analysis_subsections/11_analysis_metric_formulations}
\input{latex/0_sections/5_analysis_subsections/12_analysis_metric_sensitivity_question_num}
\input{latex/0_sections/5_analysis_subsections/13_analysis_metric_sensitivity_setting_num}

\input{latex/0_sections/5_analysis_subsections/14_analysis_parsing_errors}
\input{latex/0_sections/5_analysis_subsections/15_analysis_large_qwen}


\section{Additional Figures and Figure Details}
\input{latex/3_figure_tex/appendix_full_parameter_volatility_by_category}
\input{latex/3_figure_tex/appendix_all_different_experimental_outcomes}

\subsection{Heatmap Label Keys}
\label{app:heatmap-label-keys}
The following is the key for the column labels for Figure~\ref{fig:stratified_volatility}:
\begin{itemize}
    \item \textbf{bb} = barebones, \textbf{ya} = youare, \textbf{yak} = youareknowledge, \textbf{in} = identitynature, \textbf{ja} = justanswer, \textbf{ja-p} = justanswer-persona, \textbf{e} = explain, \textbf{e-p} = explain-persona
\end{itemize}

The following is the key for the heatmap labels for Figure~\ref{fig:dataset_and_persona_volatility}:
\begin{itemize}
    \item Dataset: \textbf{All Cat} = All Categories, \textbf{History} = History, \textbf{Pol/Law} = Politics and Law, \textbf{Soc Sci} = Social Science, \textbf{Bus/Acc/Econ} = Business/Accounting and Economics, \textbf{EECS} = EECS, \textbf{Math} = Math, \textbf{HHM} = Human Health and Medicine, \textbf{Culture} = Cultural Awareness and Understanding, \textbf{Com Sen} = Commonsense Reasoning/Social Awareness
    \item Persona: \textbf{All Cat} = All Categories, \textbf{Gender} = Gender, \textbf{Sexuality} = Sexuality, \textbf{Race/Ethn} = Race/Ethnicity, \textbf{Age} = Age, \textbf{Disabilities} = Disabilities, \textbf{Rel Beliefs} = Religious Beliefs, \textbf{Pol Beliefs} = Political Beliefs, \textbf{Occup Roles} = Occupational Roles, \textbf{EECS/Math} = Occupational roles related to the EECS/Math categories in $\mathcal{D}$, \textbf{Hist./P\&L} = Occupational roles related to the History/Politics and Law categories in $\mathcal{D}$, \textbf{Nat. Sci./HHM} = Occupational roles related to the Natural Science/Human Health and Medicine categories in $\mathcal{D}$, \textbf{Baselines} = Baselines 
\end{itemize}

%% file: latex/1_tables/prompt_table.tex
\begin{table*}[]
\tiny
\centering
\begin{tabular}{@{}cccl@{}}
\toprule
\textbf{\begin{tabular}[c]{@{}c@{}}Persona Prompt\\ Format (PPF)\\ Name\end{tabular}} & \textbf{Prompt}                                                                                                                                                                                 & \textbf{\begin{tabular}[c]{@{}c@{}}Task Prompt\\ Format (TPF)\\ Name\end{tabular}} & \multicolumn{1}{c}{\textbf{Prompt}}                                                                                                                                                                                                                                                                                           \\ \midrule
\multicolumn{1}{c|}{barebones}                                                        & \multicolumn{1}{c|}{"\{persona\}"}                                                                                                                                                              & \multicolumn{1}{c|}{justanswer}                                                    & \begin{tabular}[c]{@{}l@{}}"The following is a multiple choice question with answer\\ options. Reply with only the option number in this \\ format: {[}X{]}. \{question\} Answer:"\end{tabular}                                                                                                                               \\ \midrule
\multicolumn{1}{c|}{youare}                                                           & \multicolumn{1}{c|}{"You are \{persona\}."~\cite{zheng-jurgens-etal-2024-helpful}}                                                                                                                                                     & \multicolumn{1}{c|}{explain}                                               & \begin{tabular}[c]{@{}l@{}}"The following is a multiple choice question with answer\\ options. Answer the question and make sure to explain\\ your reasoning. In the last sentence, reply with only the\\ option number in this format: {[}X{]}. \{question\} Answer:"\end{tabular}                                           \\ \midrule
\multicolumn{1}{c|}{youareknowledge}                                                  & \multicolumn{1}{c|}{\begin{tabular}[c]{@{}c@{}}"You are \{persona\}. Your responses should\\ closely mirror the knowledge and \\ abilities of this persona."~\cite{gupta2024personabias}\end{tabular}}                      & \multicolumn{1}{c|}{justanswer-persona}                                            & \begin{tabular}[c]{@{}l@{}}"The following is a multiple choice question with answer\\ options. Consider your persona as you answer. Reply with\\ only the option number in this format: {[}X{]}. \{question\}\\ Answer:"\end{tabular}                                                                                         \\ \midrule
\multicolumn{1}{c|}{identitynature}                                                   & \multicolumn{1}{c|}{\begin{tabular}[c]{@{}c@{}}"Adopt the identity of \{persona\}. Answer the\\ questions while staying in strict accordance\\ with the nature of this identity."~\cite{gupta2024personabias}\end{tabular}} & \multicolumn{1}{c|}{explain-persona}                                               & \begin{tabular}[c]{@{}l@{}}"The following is a multiple choice question with answer\\ options. Answer the question and make sure to explain\\ your reasoning, considering your persona as you answer.\\ In the last sentence, reply with only the option number in\\ this format: {[}X{]}. \{question\} Answer:"\end{tabular}
\end{tabular}
\caption{The various prompt formats used, relative to the task prompt format (TPF) and persona prompt format (PPF) experiment configuration parameters. We substitute \{persona\} with a persona string in PPF and substitute \{question\} with a formatted question string in TPF.}
\label{tab:prompt-table}
\end{table*}


%% file: latex/1_tables/full_dataset_table.tex
\begin{table*}[]
\tiny
\centering
\begin{tabular}{@{}cccl@{}}
\toprule
\textbf{Dataset Category}                                   & \textbf{\begin{tabular}[c]{@{}c@{}}Total \#\\ Questions\end{tabular}} & \textbf{\begin{tabular}[c]{@{}c@{}}Total \#\\ Labels\end{tabular}} & \multicolumn{1}{c}{\textbf{Sub-Labels}}                                                                                                                                                                          \\ \midrule
\multicolumn{1}{c|}{History}                                & \multicolumn{1}{c|}{500}                                              & \multicolumn{1}{c|}{4}                                             & 'prehistory', 'high\_school\_us\_history', 'high\_school\_world\_history', 'high\_school\_european\_history'                                                                                                             \\ \midrule
\multicolumn{1}{c|}{Politics and Law}                       & \multicolumn{1}{c|}{500}                                              & \multicolumn{1}{c|}{5}                                             & \begin{tabular}[c]{@{}l@{}}'us\_foreign\_policy', 'professional\_law', 'jurisprudence', 'high\_school\_government\_and\_politics',\\ 'international\_law'\end{tabular}                                                   \\ \midrule
\multicolumn{1}{c|}{Social Science}                         & \multicolumn{1}{c|}{497}                                              & \multicolumn{1}{c|}{7}                                             & \begin{tabular}[c]{@{}l@{}}'sociology', 'philosophy', 'high\_school\_psychology', 'moral\_disputes', 'moral\_scenarios', 'logical\_fallacies',\\ 'professional\_psychology'\end{tabular}                                 \\ \midrule
\multicolumn{1}{c|}{Business/Accounting and Economics}      & \multicolumn{1}{c|}{496}                                              & \multicolumn{1}{c|}{8}                                             & \begin{tabular}[c]{@{}l@{}}'professional\_accounting', 'public\_relations', 'econometrics', 'business\_ethics', 'management',\\ 'high\_school\_microeconomics', 'marketing', 'high\_school\_macroeconomics'\end{tabular} \\ \midrule
\multicolumn{1}{c|}{EECS}                                   & \multicolumn{1}{c|}{500}                                              & \multicolumn{1}{c|}{5}                                             & \begin{tabular}[c]{@{}l@{}}'high\_school\_computer\_science', 'machine\_learning', 'college\_computer\_science', 'electrical\_engineering',\\ 'computer\_security'\end{tabular}                                          \\ \midrule
\multicolumn{1}{c|}{Math}                                   & \multicolumn{1}{c|}{498}                                              & \multicolumn{1}{c|}{6}                                             & \begin{tabular}[c]{@{}l@{}}'high\_school\_statistics', 'formal\_logic', 'college\_mathematics', 'elementary\_mathematics', 'abstract\_algebra',\\ 'high\_school\_mathematics'\end{tabular}                               \\ \midrule
\multicolumn{1}{c|}{Natural Science}                        & \multicolumn{1}{c|}{496}                                              & \multicolumn{1}{c|}{8}                                             & \begin{tabular}[c]{@{}l@{}}'conceptual\_physics', 'high\_school\_chemistry', 'college\_physics', 'astronomy', 'college\_biology',\\ 'high\_school\_physics', 'high\_school\_biology', 'college\_chemistry'\end{tabular}  \\ \midrule
\multicolumn{1}{c|}{Human Health and Medicine}              & \multicolumn{1}{c|}{504}                                              & \multicolumn{1}{c|}{9}                                             & \begin{tabular}[c]{@{}l@{}}'college\_medicine', 'anatomy', 'nutrition', 'medical\_genetics', 'clinical\_knowledge', 'human\_sexuality',\\ 'human\_aging', 'professional\_medicine', 'virology'\end{tabular}              \\ \midrule
\multicolumn{1}{c|}{Cultural Awareness and Understanding}   & \multicolumn{1}{c|}{525}                                              & \multicolumn{1}{c|}{75}                                            & All countries in NormAd-Eti labels~\cite{rao-etal-2025-normad}                                                                                                                                                                                          \\ \midrule
\multicolumn{1}{c|}{Commonsense Reasoning/Social Awareness} & \multicolumn{1}{c|}{497}                                              & \multicolumn{1}{c|}{7}                                             & 'motivations', 'effects', 'misc', 'reactions', 'descriptions', 'needs', 'wants'~\cite{sap-etal-2019-social-socialiqa-siqa}                                                                                                                                          \\ \bottomrule
\end{tabular}
\caption{The total composition of the MCQA dataset used in experimentation, including the question categorization.}
\label{tab:full-dataset-table}
\end{table*}

%% file: latex/1_tables/full_persona_table.tex
\begin{table*}[]
\tiny
\centering
\begin{tabular}{@{}c|c|l@{}}
\toprule
\textbf{Persona Category} & \textbf{\begin{tabular}[c]{@{}c@{}}Total \#\\ Personas\end{tabular}} & \multicolumn{1}{c}{\textbf{Personas}}                                                                                                                                         \\ \midrule
Gender                    & 5                                                                    & a man, a woman, a transgender man, a transgender woman, a non-binary person                                                                                                   \\ \midrule
Sexuality                 & 3                                                                    & a straight person, a gay person, a lesbian person                                                                                                                             \\ \midrule
Race/Ethnicity            & 6                                                                    & an African person, an Asian person, a Caucasian person, a Chinese person, an Egyptian person, a Hispanic person                                                               \\ \midrule
Age                       & 3                                                                    & an adult, a young person, an old person                                                                                                                                       \\ \midrule
Disabilities              & 2                                                                    & a disabled person, an able-bodied person                                                                                                                                      \\ \midrule
Religious Beliefs         & 4                                                                    & a religious person, an Atheist person, a Christian person, a Jewish person                                                                                                    \\ \midrule
Political Beliefs         & 4                                                                    & a Democrat, a Republican, a Barack Obama supporter, a Donald Trump supporter                                                                                                  \\ \midrule
Occupational Roles        & 12                                                                   & a virtual assistant, a lawyer, a legal scholar, a historian, a doctor, a physiologist, a biologist, an engineer, a computer engineer, a computer scientist, an artist, a chef \\ \midrule
Baselines                 & 2                                                                    & a human, NO PERSONA                                                                                                                                                           \\ \bottomrule
\end{tabular}
\caption{All personas used in experimentation.}
\label{tab:full-persona-table}
\end{table*}

%% file: latex/0_sections/5_analysis_subsections/7_analysis_persona_as_parameter.tex
\subsection{Instability Due to Persona Role Variation}
\label{subsec:persona-as-probe-for-bias}

\input{latex/3_figure_tex/heatmap_persona_as_parameter}

We explore the differences in performance or instability when using different personas. While in Section~\ref{sec:results_analysis}, we explore instability due to varying experiment settings, here, we focus on instability due to varying the persona role. By treating persona as a hyperparameter that can be varied across experiment settings, we can calculate instability relative to using different personas for the same task\footnote{For example, accuracy differences when using different personas on the same task ($I_A$), or consistency of questions correct when using different personas on the same task ($I_Q$).}. This can additionally be viewed as a lens into a form of \textit{persona bias} -- different personas on this task should perform similarly, represented by a small $I_A$ and $I_Q$. We omit an analysis of $I_O$, as that requires considering several personas \textit{simultaneously} for one experiment setting, while we treat persona as a hyperparameter \textit{part of} the experiment setting. Note that none of the questions in our dataset have correctness that is persona-dependent\footnote{While question information can be related to a specific role, like biology questions relating to "a biologist", this knowledge is not \textit{exclusive} to those with that role.}.

We calculate instability when varying the persona role, both across all personas, and within specific persona subsets. Results are shown in Figure~\ref{fig:persona_as_parameter_volatility}.

Looking at $I_A$, we can observe patterns of persona bias in accuracy similar to what is seen in past work~\cite{gupta2024personabias}. The smaller Llama 1B model has more persona bias than Llama 8B, and the Qwen models have less bias overall -- we see this with the "Disabilities" category, where Llama 1B's darker color shows greater performance differences between personas in that category.

With $I_Q$, we can get a new view of persona bias via question consistency. Even in the larger models, the different persona categories have great variation in their question consistency, with some categories like Gender having less question stability than other categories like Age, despite personas in these categories not warranting different performance (e.g. "a man" and "a transgender woman" should not perform differently on this evaluation, similar to how "a young person" and "an adult" should not perform differently as well). This shows a new view of bias even for larger models, and emphasizes the distinctness of our volatility metrics. We also still find great $I_Q$ even in our more closely related persona sub-categories where a lower value would be expected, suggesting an inherently unstable persona integration mechanism in LLMs. This analysis additionally demonstrates how the usage of persona is concretely affecting the stability of model performances, rather than model performance being inherently unstable and persona usage being an insignificant factor added on top.

%% file: latex/3_figure_tex/heatmap_persona_as_parameter.tex
\begin{figure}[!h]
    \small
    \centering
    \includegraphics[width=\columnwidth]{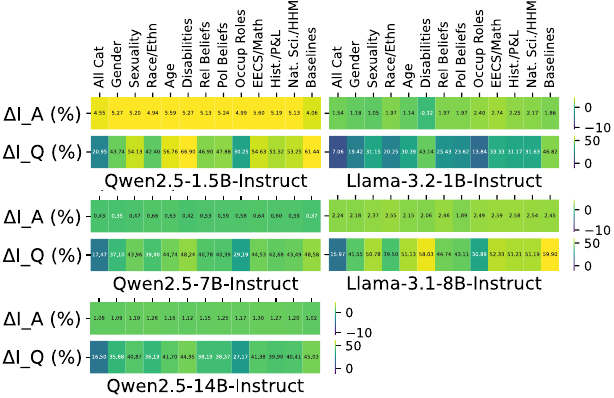}
    \caption{A figure depicting the delta hyperparameter instability when treating persona as a hyperparameter compared to the overall baseline, stratified by subsets of personas relative to the persona categorization.}
    \label{fig:persona_as_parameter_volatility}
\end{figure}

%% file: latex/0_sections/5_analysis_subsections/8_sub_occupation_category_dataset_category_analysis.tex
\subsection{Instability in Only Occupation Sub-Category-Related Dataset Categories}
\label{app:sub-category-dataset-category}

\input{latex/3_figure_tex/sub_category_and_dataset_category_analysis}

We consider Figure~\ref{fig:dataset_and_persona_volatility} for our sub-category analysis, referring to the categorization in Appendix~\ref{app:persona-details}. Due to the high similarity of the occupations in these categories in both phrasing and expertise, ideal model behavior should have \textit{less} instability within these sub-categories, compared to considering all occupations (where subject expertise is more varied) or other identity-related persona categories (where in an unbiased setting there should be no subject expertise conferred, and therefore, potentially more varied performances). However, instability considering only these sub-categories of personas mirrors other persona categories, supporting our claim that \textit{specific} personas do not strongly impact instability.

We further this analysis by evaluating how instability may vary in expertise-related domains, seeking to understand if task-aligned personas are more stable within their task than others. For this analysis, we compute the dataset category-wise instability, but only considering specific subsets of personas, aligning with the sub-categorization we discussed previously. We additionally introduce a "Misc." category of expertise-aligned personas that do not relate to any task domain -- "a virtual assistant", "a chef", and "an artist" -- to allow for more robust analysis. These results are depicted in Figure~\ref{fig:app:sub_and_data_category}. As can be seen in Figure~\ref{fig:app:sub_and_data_category}, no matter the specific persona sub-category considered, dataset domain-specific instability patterns remain the same. This supports our previous claims that the \textit{specific} persona may not impact PDG instability in MCQA

%% file: latex/3_figure_tex/sub_category_and_dataset_category_analysis.tex
\begin{figure*}
    \centering
    \includegraphics[width=\textwidth]{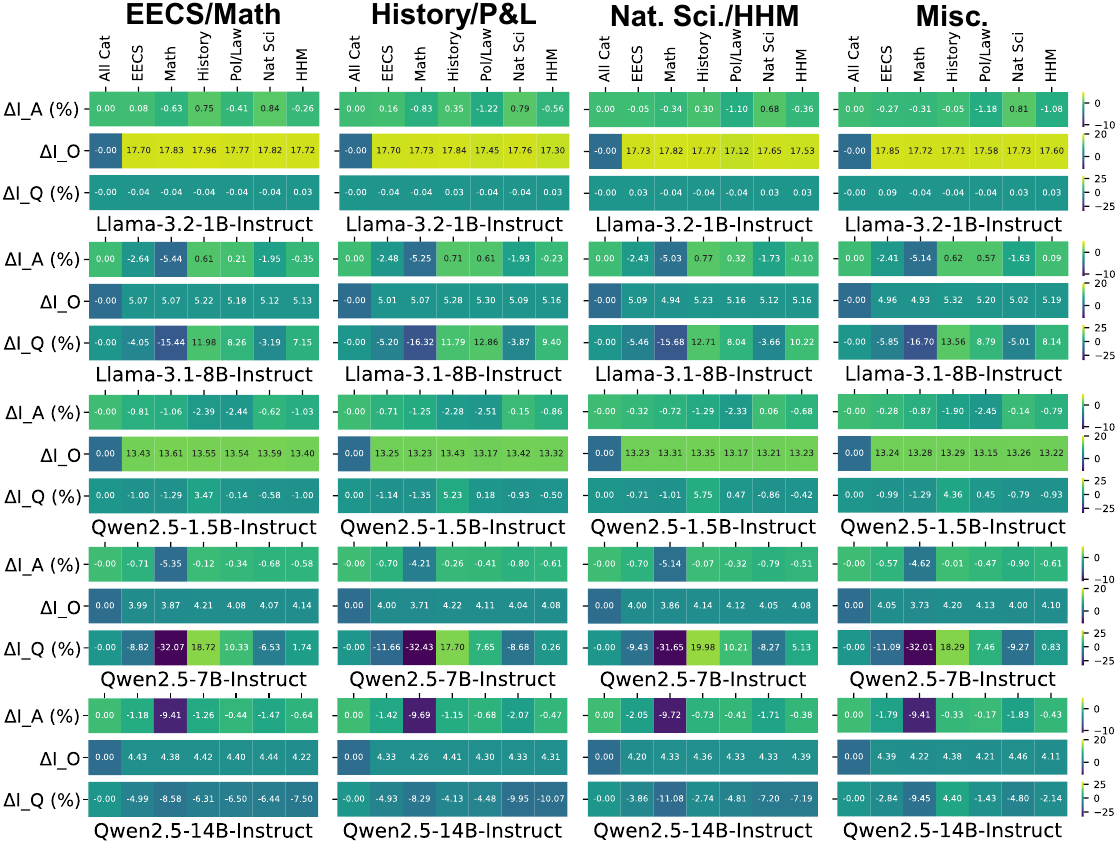}
    \caption{Instability results stratified by instability metric, by dataset category, by persona sub-category.}
    \label{fig:app:sub_and_data_category}
\end{figure*}

%% file: latex/0_sections/5_analysis_subsections/9_analysis_classifier.tex
\subsection{Instability and Predicting the Best Persona}
\label{subsec:classifier_performance}
\input{latex/3_figure_tex/classifier_performance}

We consider the issue of predicting the best persona for a task, as defined in \citet{zheng-jurgens-etal-2024-helpful}. We employ identical methods, fine-tuning a roberta-base model and using it as a multi-label classifier for personas, with the prediction target being the 41 persona roles used throughout experimentation. We train an individual classifier on results from each experimental setting across models, leading to 48 classifiers per model. We report F1 score results in Figure~\ref{fig:app:classifier_performance}, where each point represents the F1 performance of one classifier trained on one experimental setting's results. We additionally report classifier performances stratified by the hyperparameter settings of the data they are trained on.

As can be observed in Figure~\ref{fig:app:classifier_performance}, we see little relationship between instability and predictability of persona roles for a question. Using what is previously found to be stable or unstable settings per hyperparameter, such as a $\tau=0$ and a $\tau=1$, we observe no trend between classification performance and stability of hyperparameter settings.

However, we do see a concrete pattern with classification performance and TPF -- classification performance is lower across the "justanswer" variants compared to the "explain" variants of TPF across all models (except Qwen2.5-1.5B), with this pattern being strongest in the larger models. While we cannot assume stability mediates the unpredictability of PDGs in non-text-heavy settings, we can see an effect of TPF. This could be due to the "explain" variants guiding the model to be more "faithful" and consistent with respect to their persona, in turn making their performance predictable. We encourage further investigation into this phenomenon.

%% file: latex/3_figure_tex/classifier_performance.tex
\begin{figure*}
    \centering
    \includegraphics[width=\textwidth]{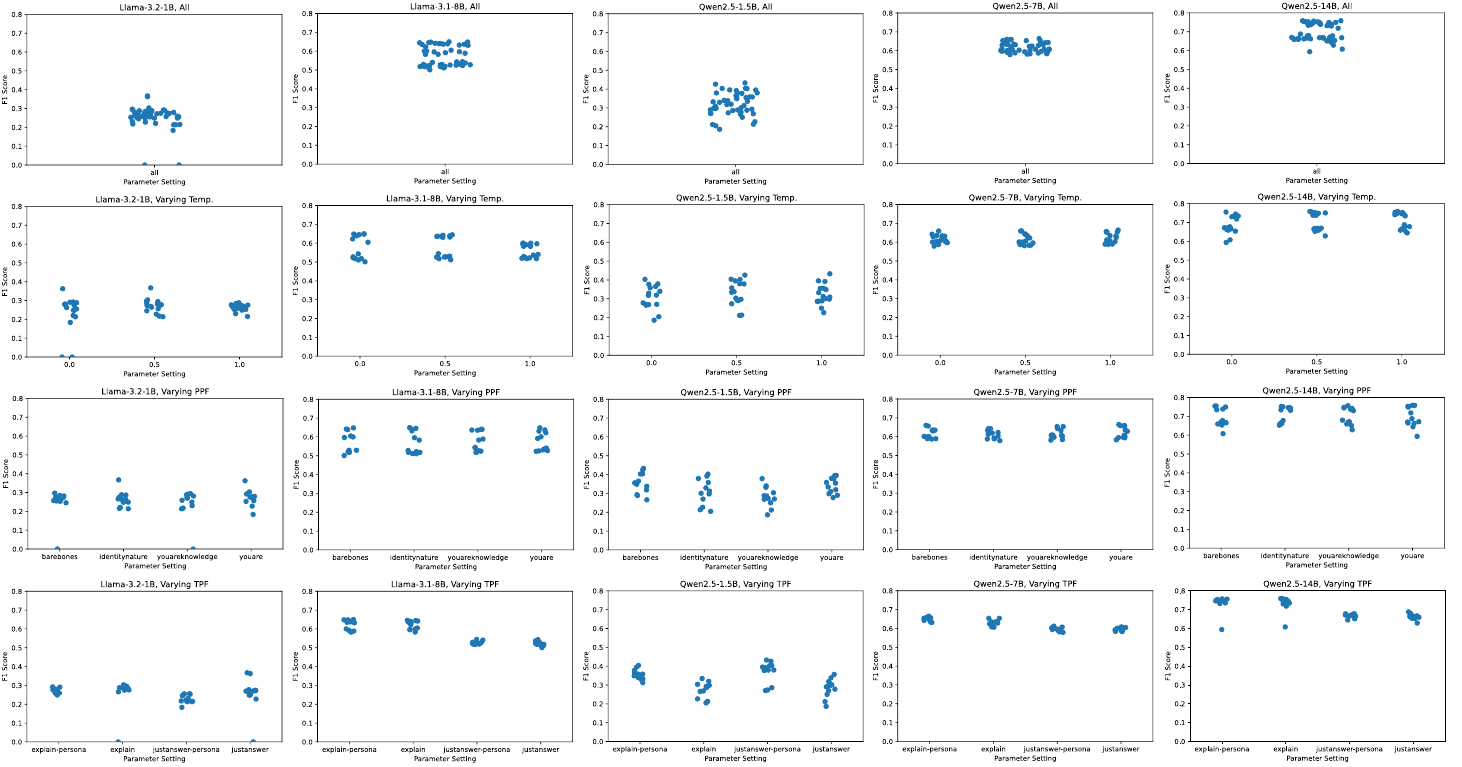}
    \caption{All classifier performances when trained across 48 experimental settings, additionally stratified by specific hyperparameter settings within the experimental settings.}
    \label{fig:app:classifier_performance}
\end{figure*}

%% file: latex/0_sections/5_analysis_subsections/10_analysis_correlation.tex
\subsection{Correlations with Existing Metrics}
\label{app:metric_correlations}

To validate our metrics, we perform various correlation studies, finding the Pearson Correlation between our metrics and selected metrics from past work. We calculate this between our metrics and similar metrics from literature or reasonable correlates that we explain. We calculate these metrics for sets of experimental settings, where each set is defined by settings where one parameter is kept static (such as a set of all possible experimental settings with $\tau$ = 0.0).
Specifically, we compare:

\begin{itemize}
\item $I_A$ and the accuracy spread (max - min) across all experiment settings of total dataset accuracy averaged across all personas for each setting. We compare against this metric from prior literature~\cite{sclar2023quantifying-spurious-prompt-features} as they investigated robustness to differing prompt features with respect to performance, similar to our investigations into performance instability across experiment settings.

\item $I_O$, without the multiplication of the average parity for a model, and the number of different best-performing personas on the total dataset across all experimental settings. We choose the number of best-performing personas as that is a direct experimental outcome (the best persona for a task) that we seek to capture with our metric. Due to reasons relating to initial persona parity for a model as discussed in the paper, we argue that the number of best-performing personas across experiment settings would also be affected by the same quality. Because of this, we remove the regularization term that would control for this quality in order to have more equivalent comparison.
        
\item $I_Q$ and the Fleiss’ Kappa~\cite{fleisskappa} agreement of the binary vectors of question correctness for the total dataset between experiment settings for one persona, averaged across all personas. Here, we treat separate experiment settings for a persona as different “annotators”. This is chosen as agreement between annotators is used in literature as a measure of instance-wise consistency/stability across annotators. Note that a higher agreement rate corresponds to a lower $I_Q$ (or a positive benefit from the baseline), so we would expect to observe a strong positive correlation if these two measurements are consistent, in contrast to the other pairs.
\end{itemize}

We report the p-value and r of this correlation, averaged across all models, for each of our metrics, in Table~\ref{tab:correlation_with_other_metrics}. We observe strong correlations.

\input{latex/1_tables/correlation_with_other_metrics}

%% file: latex/1_tables/correlation_with_other_metrics.tex
\begin{table*}[]
\small
\centering
\begin{tabular}{@{}c|cc|cc|cc@{}}
\toprule
\textbf{\begin{tabular}[c]{@{}c@{}}Instability\\ Metric Pairs\end{tabular}} & \multicolumn{2}{c|}{$I_A \times$ \textbackslash{}textbf\{Spread\}} & \multicolumn{2}{c|}{$I_O \times$ \textbackslash{}textbf\{Num. Best.\}} & \multicolumn{2}{c}{$I_Q \times$ \textbackslash{}textbf\{Fleiss' Kappa\}} \\ \midrule
\textit{Models}                                                             & p                               & r                                & p                                 & r                                  & p                                   & r                                  \\ \midrule
Llama-3.2-1B                                                                & 0.002                           & -0.811                           & 0.049                             & -0.603                             & 0.000                               & 0.948                              \\
Llama-3.1-8B                                                                & 0.005                           & -0.779                           & 0.111                             & -0.508                             & 0.000                               & 0.954                              \\
Qwen2.5-1.5B                                                                & 0.029                           & -0.653                           & 0.017                             & -0.696                             & 0.000                               & 0.986                              \\
Qwen2.5-7B                                                                  & 0.019                           & -0.688                           & 0.040                             & -0.624                             & 0.000                               & 0.990                              \\
Qwen2.5-14B                                                                 & 0.000                           & -0.938                           & 0.005                             & -0.781                             & 0.000                               & 0.990                              \\ \midrule
\textit{Mean}                                                               & 0.011                           & -0.774                           & 0.044                             & -0.642                             & 0.000                               & 0.974                              \\ \bottomrule
\end{tabular}
\caption{Depicting the p-value and r of correlations between each pair of metrics across all models.}
\label{tab:correlation_with_other_metrics}
\end{table*}

%% file: latex/0_sections/5_analysis_subsections/11_analysis_metric_formulations.tex
\subsection{Alternative Formulations of $I_O$}
\label{app:i_o_formulations}
\input{latex/3_figure_tex/combined_different_formulations}

We experimented with four formulations of $I_O$:
\begin{itemize}
    \item A non-weighted standard deviation of ranks
    \item A non-weighted standard deviation of ranks, with additional multiplication of a model normalization term (as previously defined)
    \item A weighted standard deviation of ranks, weighted by the persona parity of each experimental setting
    \item A weighted standard deviation of ranks, weighted by the persona parity of each experimental setting, with additional multiplication of a model normalization term (as previously defined)
\end{itemize}

As we consider that ranking differences from states of parity "mean less" than those coming from states of a lack of parity (i.e. ranking fluctuations when all personas perform equally means less than ranking fluctuations when all personas perform very differently to one another), we want to weight a given ranking by the parity of the setting it comes from. With this, rankings that come from settings with great parity are weighted less in the final calculation. From this, we focus on the latter two formulations, using a weighted standard deviation with the parity of the experimental settings being used as the weights.

We then investigate the usage of the final model normalization term. Without this final model normalization term, we have the potential for the weighted standard deviation to be "washed out" -- in cases where, across experimental settings, models have equivalent levels of high parity, all ranks will be weighted the same, potentially leading to the final $I_O$ score being inflated. To account for this, we add an additional model normalization term, integrating overall parity a model might have. This way, should the weighted standard deviation be "washed out", we are still able to integrate a view of model parity, which we deem as valuable for reasons previously stated.

With this, we wanted to evaluate whether this final model normalization term would "dominate" the final score. Thus, we investigate patterns of solely the model normalization term, and of the $I_O$ implementations -- with this, we seek to judge whether the patterns of the formulation \textit{with} the model normalization term completely follow those of solely the model normalization term, implying the model normalization term may be "too strong". We report heatmaps of the results in Figure~\ref{fig:app:i_o_formulations}.

From Figure~\ref{fig:app:i_o_formulations}, we can observe that, while A (solely the model normalization term) has similar patterns to B (weighted standard deviation with the model normalization term), they are not identical. We note key regions where they differ, such as $I_O$ for temperature across all models. From this, we go forward with the final formulation of $I_O$, calculating a weighted standard deviation of ranks across experimental settings, weighted by the parity of each setting, with the addition of a final model normalization term.

%% file: latex/3_figure_tex/combined_different_formulations.tex
\begin{figure*}
    \centering
    \includegraphics[width=\textwidth]{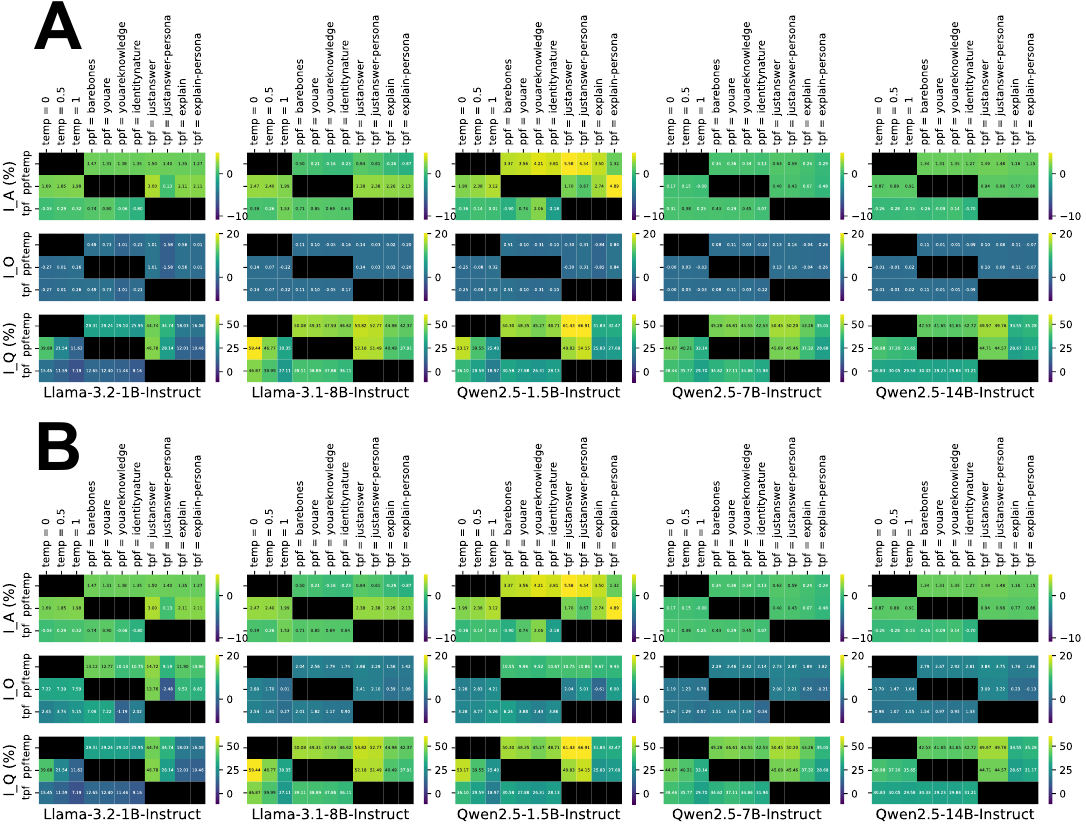}
    \caption{Comparing instability results across different formulations of $I_O$.}
    \label{fig:app:i_o_formulations}
\end{figure*}

%% file: latex/0_sections/5_analysis_subsections/12_analysis_metric_sensitivity_question_num.tex
\subsection{Analyzing Metric Sensitivity to Question Count}
\label{app:sensitivity-question-count}
\input{latex/1_tables/sensitivity_to_question_count}

We analyze the impact of the question set size on our metric results. For this analysis, we compute the baseline metrics (as in Table~\ref{tab:model_baselines}) using a randomly selected subset of ~75\% and ~50\% of the question set, balanced across dataset categories. We provide results in Table~\ref{app:sensitivity-question-count}, additionally with the current baseline results for comparison.

As can be seen in Table~\ref{app:sensitivity-question-count}, our metrics slightly vary with respect to the size of the question set. This aligns with our expectations; with a reduced number of questions, accuracy percentage differences are more substantial, reasonably affecting $I_A$ and $I_O$. Additionally, with a reduced question set, differences in the set of shared questions are similarly more substantial, affecting $I_Q$. While results vary slightly, it can be noted that they still stabilize around a similar value across dataset size/metrics across models, maintaining features such as the relative order of models for each instability metric, and the relative difference between models for each instability metric.  To summarize: as the question set size changes, the metrics vary slightly but still follow the same trend across models and experiment settings.

%% file: latex/1_tables/sensitivity_to_question_count.tex
\begin{table*}[]
\small
\centering
\begin{tabular}{@{}l|ccc|ccc|ccc@{}}
\toprule
\textbf{}                & \multicolumn{3}{c|}{\textbf{\begin{tabular}[c]{@{}c@{}}All Questions Per\\ Dataset Category (100\%)\end{tabular}}} & \multicolumn{3}{c|}{\textbf{\begin{tabular}[c]{@{}c@{}}\#=375 Per Dataset\\ Category ($\sim$75\%)\end{tabular}}} & \multicolumn{3}{c}{\textbf{\begin{tabular}[c]{@{}c@{}}\#=250 Per Dataset\\ Category ($\sim$50\%)\end{tabular}}} \\ \midrule
\textit{Models / Metric} & \textbf{$I_A$}                       & \textbf{$I_O$}                       & \textbf{$I_Q$}                       & \textbf{$I_A$}                      & \textbf{$I_O$}                      & \textbf{$I_Q$}                      & \textbf{$I_A$}                      & \textbf{$I_O$}                      & \textbf{$I_Q$}                     \\ \midrule
Llama-3.2-1B             & 3.697                                & 18.791                               & 99.960                               & 3.684                               & 19.379                              & 99.952                              & 3.715                               & 20.311                              & 99.955                             \\
Llama-3.1-8B             & 2.999                                & 5.784                                & 78.293                               & 2.980                               & 6.276                               & 78.041                              & 3.163                               & 7.139                               & 78.678                             \\
Qwen2.5-1.5B             & 6.586                                & 14.305                               & 98.576                               & 6.714                               & 15.058                              & 98.592                              & 6.660                               & 15.291                              & 98.633                             \\
Qwen2.5-7B               & 0.972                                & 4.539                                & 57.688                               & 0.963                               & 4.855                               & 57.500                              & 1.074                               & 5.702                               & 57.893                             \\
Qwen2.5-14B              & 1.623                                & 4.812                                & 52.828                               & 1.671                               & 5.278                               & 52.739                              & 1.647                               & 5.746                               & 52.451                             \\ \bottomrule
\end{tabular}
\caption{Instability metric results, stratified by the number of questions considered per dataset category. ``\#" denotes the number of questions considered for each category.}
\label{tab:sensitivity_to_question_count}
\end{table*}

%% file: latex/0_sections/5_analysis_subsections/13_analysis_metric_sensitivity_setting_num.tex
\subsection{Analyzing Metric Sensitivity to Experiment Setting Count}
\label{app:sensitivity-setting-count}
\input{latex/1_tables/sensitivity_to_experiment_count}

We analyze the impact of the number of experiment settings on our metric results. For this analysis, we compute the baseline metrics using a reduced subset of experiment settings, considering only experiment settings across all temperature settings with TPF = {justanswer, explain, explain-persona} and PPF = {barebones, youareknowledge, identitynature} ("Subset A", 27 settings), and experiment settings with TPF = {justanswer, explain-persona} and PPF = {barebones, identitynature} ("Subset B", 12 settings). We provide results in Table~\ref{tab:sensitivity_to_setting_count}, additionally with the current baseline results for comparison.

As can be seen in Table~\ref{tab:sensitivity_to_setting_count}, our metrics vary with respect to the size of the experiment setting set, similar to that of the question set size. This aligns with our expectations. Similar to how, with other measures of spread like classic standard deviation, samples from a population of gradually increasing sizes approach an approximate population value, we see a similar pattern with our metrics. We maintain features such as the relative order of models for each instability metric. To summarize: as the experiment setting set size changes, the metrics vary slightly but approach similar values and still follow the same trend across models and experiment settings.

Additionally, it is important to note that these metric calculations are, of course, dependent on the composition of the sample, like with other measures of spread. While here, the subsets of experiment settings attempt to maintain an equal count of low/high instability hyperparameter settings, differing subsets may produce considerably different results (such as a subset of only low instability hyperparameter settings).

%% file: latex/1_tables/sensitivity_to_experiment_count.tex
\begin{table*}[]
\small
\centering
\begin{tabular}{@{}l|ccc|ccc|ccc@{}}
\toprule
\textbf{}                & \multicolumn{3}{c|}{\textbf{\begin{tabular}[c]{@{}c@{}}All Experiment\\ Settings (\#=48)\end{tabular}}} & \multicolumn{3}{c|}{\textbf{Subset A (\#=27)}}   & \multicolumn{3}{c}{\textbf{Subset B (\#=12)}}    \\ \midrule
\textit{Models / Metric} & \textbf{$I_A$}                    & \textbf{$I_O$}                   & \textbf{$I_Q$}                   & \textbf{$I_A$} & \textbf{$I_O$} & \textbf{$I_Q$} & \textbf{$I_A$} & \textbf{$I_O$} & \textbf{$I_Q$} \\ \midrule
Llama-3.2-1B             & 3.697                             & 18.791                           & 99.960                           & 3.366          & 13.276         & 99.845         & 3.829          & 11.155         & 98.271         \\
Llama-3.1-8B             & 2.999                             & 5.784                            & 78.293                           & 3.121          & 6.126          & 72.863         & 3.210          & 5.528          & 61.083         \\
Qwen2.5-1.5B             & 6.586                             & 14.305                           & 98.576                           & 5.618          & 15.022         & 97.505         & 3.990          & 11.818         & 89.042         \\
Qwen2.5-7B               & 0.972                             & 4.539                            & 57.688                           & 1.070          & 5.132          & 54.071         & 1.303          & 5.112          & 43.860         \\
Qwen2.5-14B              & 1.623                             & 4.812                            & 52.828                           & 1.559          & 5.141          & 47.740         & 1.925          & 3.966          & 32.061         \\ \bottomrule
\end{tabular}
\caption{Instability metric results, stratified by the number of experiment settings considered. ``\#" denotes the number of experiment settings for a given subset. "Subset A" represents all temperature settings with TPF = \{justanswer, explain, explain-persona\} and PPF = \{barebones, youareknowledge, identitynature\}. "Subset B" represents all temperature settings with TPF = \{justanswer, explain-persona\} and PPF = \{barebones, identitynature\}.}
\label{tab:sensitivity_to_setting_count}
\end{table*}

%% file: latex/0_sections/5_analysis_subsections/14_analysis_parsing_errors.tex
\subsection{Parsing Errors Across Hyperparameter Settings}
\label{app:parsing_error_analysis}
\input{latex/1_tables/parsing_error_by_hyperparameter_setting}
\input{latex/1_tables/incorrectness_by_hyperparameter_setting}

We consider the percent of parsing errors to be the percent of questions marked incorrect due to an unparseable model response for a specific experiment setting and persona combination.
We provide a table of parsing error fraction separated by model and by experiment setting hyperparameter value in Table~\ref{tab:parsing_errors}. For each experiment setting hyperparameter value, we average the fraction of parsing errors across all experiment settings with that hyperparameter value for that specific model. For example, for Qwen2.5-1.5B and TPF = bb, we average the fraction of parsing errors for Qwen2.5-1.5B-Instruct across all experiments with TPF = barebones. We additionally provide a similar table, but with the fraction of incorrect questions (inclusive of incorrect answers and unparseable answers) along similar strata in Table~\ref{tab:incorrectness}.

We can note slight variations in parsing errors across hyperparameter settings (with smaller variations in larger models, as expected), with a few notable cases (such as Qwen2.5-1.5B comparing PPF = {bb, ya} to PPF = {yak, in}). We can note that, even in these notable cases, by comparing the total incorrect fraction to that of parsing errors, the difference in parsing errors does not account for that of the incorrect questions. In summary, while we note parsing errors, these errors are not enough to explain the noted performance differences that influence our final instability metrics.

%% file: latex/1_tables/parsing_error_by_hyperparameter_setting.tex
\begin{table*}[]
\tiny
\centering
\begin{tabular}{@{}c|c|ccc|cccc|cccc@{}}
\toprule
\textit{Model / Parsing Error} & \textbf{Overall} & \textbf{\begin{tabular}[c]{@{}c@{}}Temp =\\ 0.0\end{tabular}} & \textbf{\begin{tabular}[c]{@{}c@{}}Temp =\\ 0.5\end{tabular}} & \textbf{\begin{tabular}[c]{@{}c@{}}Temp =\\ 1.0\end{tabular}} & \textbf{\begin{tabular}[c]{@{}c@{}}PPF =\\ bb\end{tabular}} & \textbf{\begin{tabular}[c]{@{}c@{}}PPF =\\ ya\end{tabular}} & \textbf{\begin{tabular}[c]{@{}c@{}}PPF =\\ yak\end{tabular}} & \textbf{\begin{tabular}[c]{@{}c@{}}PPF =\\ in\end{tabular}} & \textbf{\begin{tabular}[c]{@{}c@{}}TPF =\\ ja\end{tabular}} & \textbf{\begin{tabular}[c]{@{}c@{}}TPF =\\ ja-p\end{tabular}} & \textbf{\begin{tabular}[c]{@{}c@{}}TPF =\\ e\end{tabular}} & \textbf{\begin{tabular}[c]{@{}c@{}}TPF =\\ e-p\end{tabular}} \\ \midrule
Llama-3.2-1B                   & 0.090            & 0.086               & 0.086               & 0.097               & 0.088             & 0.076             & 0.092              & 0.102             & 0.010             & 0.174               & 0.069            & 0.106              \\ 
Llama-3.1-8B                   & 0.015            & 0.008               & 0.008               & 0.030               & 0.014             & 0.014             & 0.016              & 0.017             & 0.004             & 0.005               & 0.025            & 0.028              \\ 
Qwen2.5-1.5B                   & 0.189            & 0.182               & 0.184               & 0.200               & 0.132             & 0.161             & 0.255              & 0.205             & 0.236             & 0.099               & 0.272            & 0.148              \\ 
Qwen2.5-7B                     & 0.013            & 0.012               & 0.012               & 0.016               & 0.010             & 0.006             & 0.014              & 0.022             & 0.001             & 0.000               & 0.020            & 0.032              \\ 
Qwen2.5-14B                    & 0.001            & 0.001               & 0.001               & 0.001               & 0.001             & 0.001             & 0.001              & 0.002             & 0.001             & 0.001               & 0.002            & 0.002              \\ \bottomrule
\end{tabular}
\caption{The percent of questions marked incorrect due to parsing errors across all models. Results are stratified by hyperparameter setting, where results for a specific hyperparameter setting represent the average across all experiment settings with that specific hyperparameter setting.}
\label{tab:parsing_errors}
\end{table*}

%% file: latex/1_tables/incorrectness_by_hyperparameter_setting.tex
\begin{table*}[]
\tiny
\centering
\begin{tabular}{@{}c|c|ccc|cccc|cccc@{}}
\toprule
\textit{Model / Parsing Error} & \textbf{Overall} & \textbf{\begin{tabular}[c]{@{}c@{}}Temp =\\ 0.0\end{tabular}} & \textbf{\begin{tabular}[c]{@{}c@{}}Temp =\\ 0.5\end{tabular}} & \textbf{\begin{tabular}[c]{@{}c@{}}Temp =\\ 1.0\end{tabular}} & \textbf{\begin{tabular}[c]{@{}c@{}}PPF =\\ bb\end{tabular}} & \textbf{\begin{tabular}[c]{@{}c@{}}PPF =\\ ya\end{tabular}} & \textbf{\begin{tabular}[c]{@{}c@{}}PPF =\\ yak\end{tabular}} & \textbf{\begin{tabular}[c]{@{}c@{}}PPF =\\ in\end{tabular}} & \textbf{\begin{tabular}[c]{@{}c@{}}TPF =\\ ja\end{tabular}} & \textbf{\begin{tabular}[c]{@{}c@{}}TPF =\\ ja-p\end{tabular}} & \textbf{\begin{tabular}[c]{@{}c@{}}TPF =\\ e\end{tabular}} & \textbf{\begin{tabular}[c]{@{}c@{}}TPF =\\ e-p\end{tabular}} \\ \midrule
Llama-3.2-1B                   & 0.649            & 0.635               & 0.645               & 0.667               & 0.642             & 0.641             & 0.651              & 0.661             & 0.615             & 0.668               & 0.647            & 0.665              \\ 
Llama-3.1-8B                   & 0.367            & 0.350               & 0.357               & 0.394               & 0.364             & 0.366             & 0.367              & 0.372             & 0.381             & 0.384               & 0.351            & 0.352              \\ 
Qwen2.5-1.5B                   & 0.563            & 0.546               & 0.555               & 0.588               & 0.535             & 0.549             & 0.595              & 0.573             & 0.576             & 0.504               & 0.617            & 0.554              \\ 
Qwen2.5-7B                     & 0.322            & 0.318               & 0.321               & 0.328               & 0.319             & 0.316             & 0.322              & 0.331             & 0.323             & 0.323               & 0.320            & 0.323              \\ 
Qwen2.5-14B                    & 0.263            & 0.263               & 0.263               & 0.264               & 0.259             & 0.262             & 0.265              & 0.267             & 0.274             & 0.277               & 0.255            & 0.247              \\ \bottomrule
\end{tabular}
\caption{The percent of questions marked incorrect across all models. Results are stratified by hyperparameter setting, where results for a specific hyperparameter setting represent the average across all experiment settings with that specific hyperparameter setting.}
\label{tab:incorrectness}
\end{table*}

%% file: latex/0_sections/5_analysis_subsections/15_analysis_large_qwen.tex
\subsection{Considering Implications of Larger Models}
\label{app:large_qwen_results}

We consider the results for Qwen2.5-14B-Instruct, as seen throughout the paper (and notably in Tables~\ref{tab:model_baselines} and \ref{tab:stability_and_accuracy}). Here, we can observe that Qwen2.5B-14B-Instruct breaks the model size-related trend for $I_A$ and $I_O$, exhibiting greater instability in accuracy and outcome (represented by these metrics) than Qwen2.5-7B-Instruct.

However, we do not believe this is due to greater inherent instability in the model, but rather due to performance improvements due to chain-of-thought (CoT) prompting~\cite{wei2022chain}. While smaller models are noted to largely not benefit from CoT, larger models do, creating large accuracy differentials between non-CoT and CoT experiment settings. As we consider CoT-like prompts within our task prompt formats (the "explain" and "explain-persona" variants), we would see larger accuracy differentials between settings using these prompts and those that do not, contributing to a great $I_A$ (due to CoT generally improving larger model performance) and $I_O$ (due to changes in accuracy subsequently changing inter-persona rankings).

Aside from this, we note that Qwen2.5-14B-Instruct maintains all other existing patterns with models in the same family across all of our analyses. While we do not evaluate larger models due to resource constraints, we reasonably expect that larger Qwen2.5 models, for instance, would continue to maintain these trends.

%% file: latex/3_figure_tex/appendix_full_parameter_volatility_by_category.tex
\begin{figure*}
    \centering
    \includegraphics[width=\textwidth]{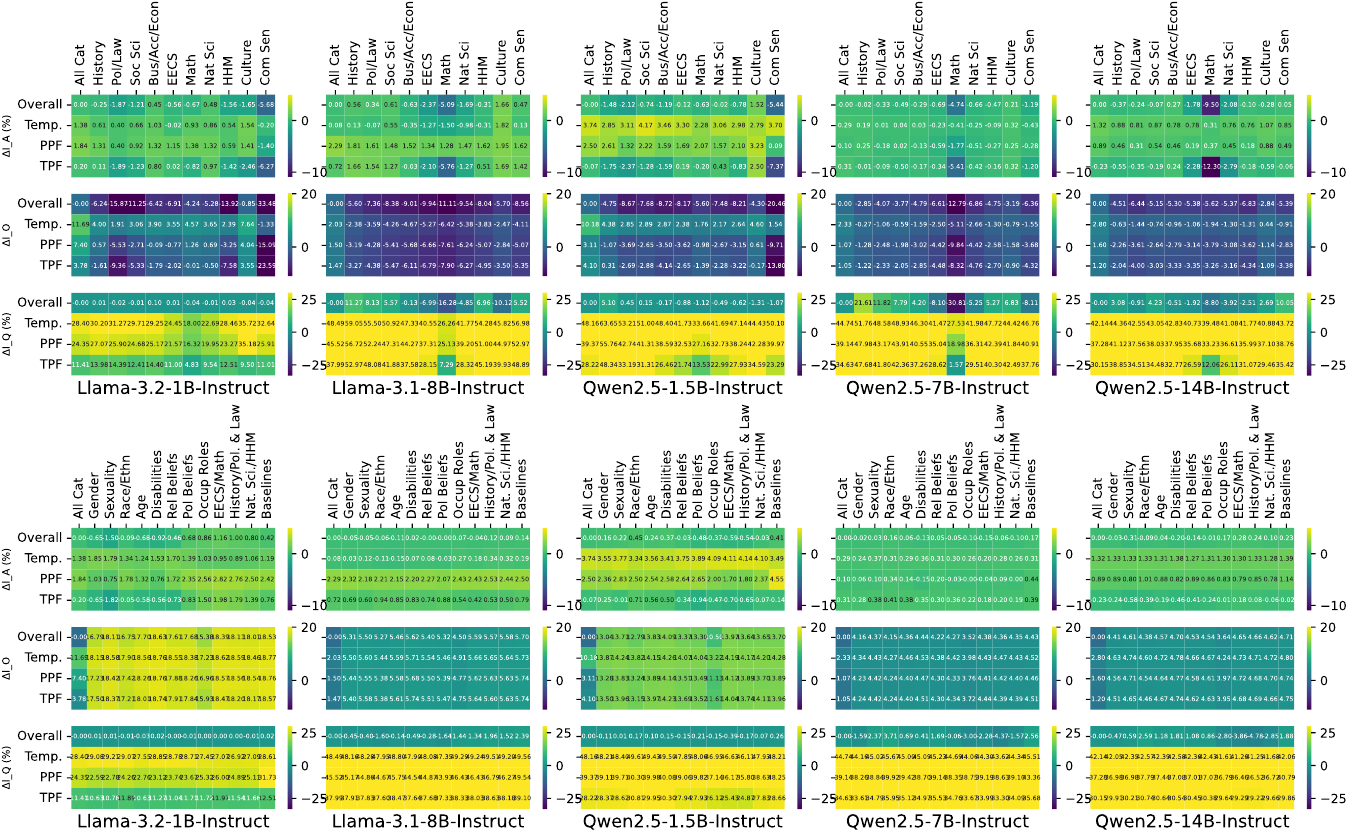}
    \caption{Hyperparameter instability results per metric, per model, stratified by dataset category and persona category.}
    \label{fig:app:full_category_analysis}
\end{figure*}

%% file: latex/3_figure_tex/appendix_all_different_experimental_outcomes.tex
\begin{figure*}
    \centering
    \includegraphics[width=\textwidth]{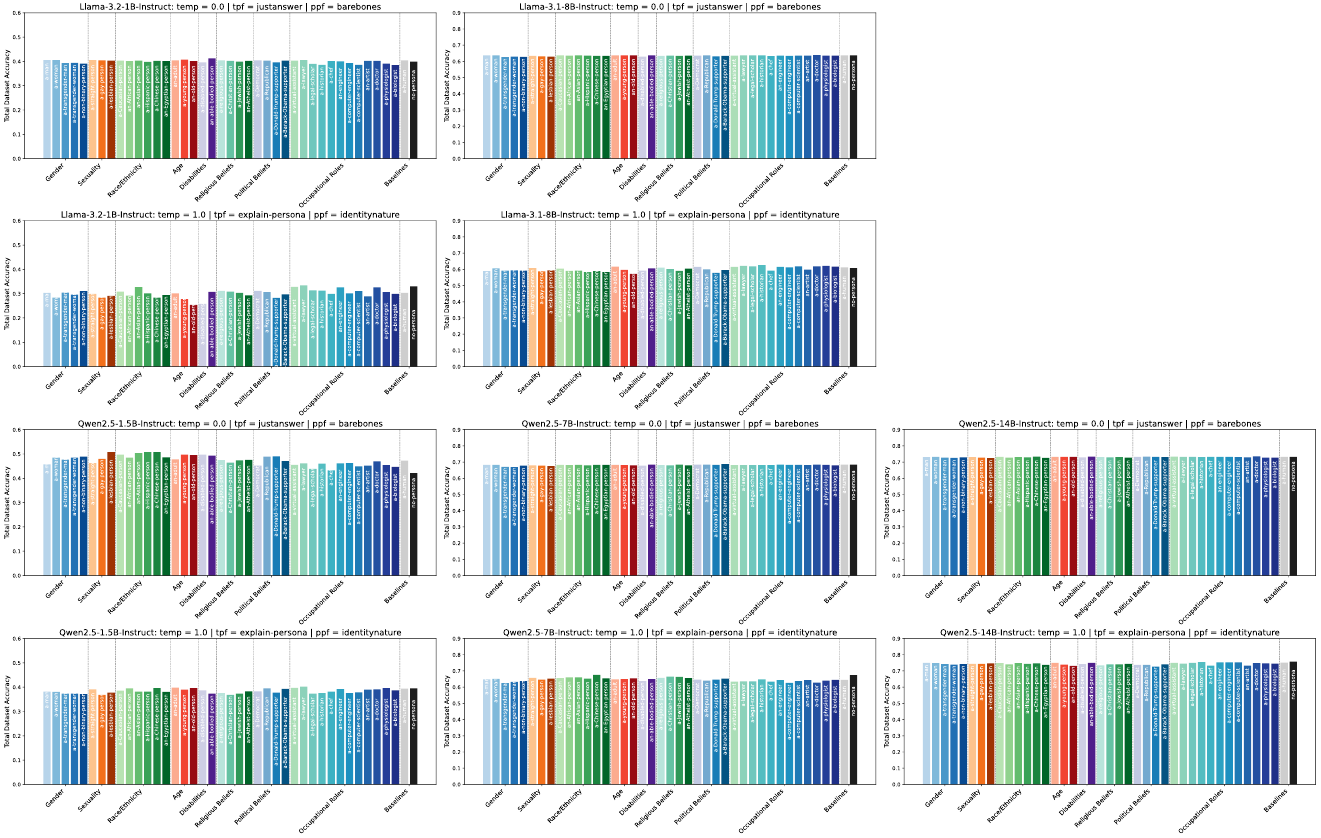}
    \caption{All inter-persona performance distributions between the "unstable" and "stable" experiment settings, as identified using our metrics, across all models.}
    \label{fig:app:all_experiment_outcomes}
\end{figure*}